\documentclass{article}

\usepackage[preprint,nonatbib]{neurips_2025}

\usepackage{hyperref}
\usepackage[utf8]{inputenc} 
\usepackage[T1]{fontenc}    
\usepackage{amsmath}
\usepackage[capitalize]{cleveref}
\usepackage{url}            
\usepackage{booktabs}       
\usepackage{amsfonts}       
\usepackage{nicefrac}       
\usepackage{microtype}      
\usepackage{xcolor}         
\usepackage{multirow}  
\usepackage{graphicx}
\usepackage{amssymb}
\usepackage{bbding}
\usepackage{pifont}
\usepackage{wasysym}
\usepackage{makecell} 
\usepackage{enumitem}
\usepackage{bm}

\renewcommand{\vec}[1]{\boldsymbol{#1}}

\newcommand{\one}{\mathbf{1}}

\newcommand{\Rot}{\mathbf{R}}
\newcommand{\K}{\mathbf{K}}
\newcommand{\Hb}{\mathbf{H}}

\newcommand{\z}{\mathbf{z}}

\newcommand{\Loss}{\mathcal{L}} 

\newcommand{\pose}[0]{\vec{\theta}}

\newcommand{\Lsmooth}{\Loss_{\mathrm{smooth}}}
\newcommand{\Langle}{\Loss_{\mathrm{angle}}}
\newcommand{\Lpalm}{\Loss_{\mathrm{palm}}}
\newcommand{\Lbl}{\Loss_{\mathrm{bl}}}
\newcommand{\Lja}{\Loss_{\mathrm{ja}}}
\newcommand{\Lbio}{\Loss_{\mathrm{bio}}}

\newcommand{\Lseg}{\mathcal{L}_{\mathrm{seg}}}
\newcommand{\Lmano}{\mathcal{L}_{\mathrm{mano}}}
\newcommand{\Ltool}{\mathcal{L}_{\mathrm{tool}}}

\newcommand{\Lshape}{\mathcal{L}_{\shape}}

\newcommand{\LTwoD}{\mathcal{L}_{\mathrm{2d}}}
\newcommand{\LpjTwoD}{\mathcal{L}_{\mathrm{pj2d}}}

\newcommand{\Lprior}{\mathcal{L}_{\mathrm{prior}}}

\newcommand{\LThreeD}{\mathcal{L}_{\mathrm{3d}}}

\newcommand{\Lsil}{\mathcal{L}_{\mathrm{sil}}}
\newcommand{\Lsdf}{\mathcal{L}_{\mathrm{sdf}}}
\newcommand{\Linter}{\mathcal{L}_{\mathrm{inter}}}
\newcommand{\Lpose}{\mathcal{L}_{\pose}}

\newcommand{\vid}{\mathcal{V}}

\newcommand{\Lmat}{\mathbf{L}}
\newcommand{\handvert}{\mathbf{V}}
\newcommand{\handjoints}{\mathbf{J}}
\newcommand{\skinweight}{\mathbf{S}}
\newcommand{\hand}{\bm{q}}

\newcommand{\joints}{\bm{J}}

\newcommand{\shape}{\bm{\beta}}
\newcommand{\orient}{\bm{\phi}}
\newcommand{\trans}{\bm{\tau}}
\newcommand{\pos}{\bm{{\tau}}}

\newcommand{\corient}{^{\mathrm{c}}\bm{\phi}}
\newcommand{\ctrans}{\bm{^{\mathrm{c}}\mathrm{\tau}}}

\newcommand{\worient}{^{\mathrm{w}}\bm{\phi}}
\newcommand{\wtrans}{\bm{^{\mathrm{w}}\mathrm{\tau}}}

\newcommand{\whandjoints}{^{\mathrm{w}}\mathbf{J}}

\newcommand{\bones}{\mathbf{b}}
\newcommand{\curvs}{\mathbf{c}}
\newcommand{\dang}{\mathbf{d}}
\newcommand{\conf}{\mathbf{C}}

\crefname{equation}{eq.}{eq.}
\Crefname{equation}{Eq.}{Eq.}
\crefname{theorem}{thm.}{thms.}
\Crefname{Theorem}{Thm.}{Thms.}
\crefname{conjecture}{conj.}{conjs.}
\Crefname{Conjecture}{Conj.}{Conjs.}
\crefname{proposition}{prop.}{props.}
\Crefname{proposition}{Prop.}{Props.}
\crefname{definition}{dfn.}{dfn.}
\Crefname{definition}{Dfn.}{Dfn.}
\crefname{remark}{remark}{remark}
\Crefname{Remark}{Remark}{Remark}
\Crefname{algorithm}{Alg.}{Alg.}

\newcommand{\ie}{\textit{i}.\textit{e}. }

\crefname{section}{Sec.}{Secs.}
\Crefname{section}{Sec.}{Secs.}
\crefname{equation}{Eq.}{Eqs.}
\Crefname{equation}{Eq.}{Eqs.}
\crefname{figure}{Fig.}{Figs.}
\Crefname{figure}{Fig.}{Figs.}
\crefname{table}{Tab.}{Tabs.}
\Crefname{table}{Tab.}{Tabs.}
\crefname{thm}{Thm.}{Thms.}
\Crefname{thm}{Thm.}{Thms.}
\crefname{conj}{Conj.}{Conjs.}
\Crefname{conj}{Conj.}{Conjs.}
\crefname{dfn}{Dfn.}{Dfns.}
\crefname{dfn}{Dfn.}{Dfns.}
\crefname{remark}{remark}{remarks}
\Crefname{Remark}{Remark}{Remarks}
\crefname{prop}{Prop.}{Prop.}
\Crefname{prop}{Prop.}{Prop.}
\Crefname{algorithm}{Alg.}{Alg.}
\crefname{appendix}{App.}{apps.}
\Crefname{appendix}{App.}{Apps.}
\crefname{appsec}{appendix}{appendices}
\Crefname{appsec}{Appendix}{Appendices}

\newcommand{\eg}{\textit{e}.\textit{g}. }

\renewcommand{\paragraph}[1]{{\vspace{0.3mm}\noindent \bf #1}}

\definecolor{darkgreen}{rgb}{0.0, 0.5, 0.0}

\usepackage{bmpsize}
\usepackage{adjustbox}
\usepackage{caption}
\usepackage{subcaption}

\let\oldding\ding 
\renewcommand{\ding}[1]{%
  \ifnum#1=51 \textcolor[RGB]{61,145,64}{\oldding{51}}
  \else\ifnum#1=55 \textcolor[RGB]{227,23,13}{\oldding{55}}
  \else\oldding{#1}\fi\fi
}

\usepackage{array}
\usepackage{multirow}

\usepackage{array}
\usepackage{tikz}

\newcommand{\VertDashedLine}[1]{%
  \begin{tikzpicture}[baseline=(current bounding box.north)]
    \draw[dashed, line width=0.6pt] (0,0) -- (0,-#1);
  \end{tikzpicture}%
}

\usepackage{placeins}

\usepackage{subcaption} 
\usepackage{calc}
\usepackage{tikz} 
\usepackage{xparse}
\usepackage{wrapfig}

\title{Towards Dynamic 3D Reconstruction of \\
Hand-Instrument Interaction in Ophthalmic Surgery}

\author{Ming Hu\textsuperscript{1,2,3}\thanks{Equal contribution, $^\dagger$Corresponding author}\quad 
    Zhengdi Yu\textsuperscript{4}$^{*}$\quad 
    Feilong Tang\textsuperscript{1,2,3}\quad 
    Kaiwen Chen\textsuperscript{5}\quad 
    Yulong Li\textsuperscript{3}\\\quad 
    \textbf{Imran Razzak}\textsuperscript{3}\quad 
    \textbf{Junjun He}\textsuperscript{2}\quad 
    \textbf{Tolga Birdal}\textsuperscript{4}\quad 
    \textbf{Kaijing Zhou}\textsuperscript{5}$^\dagger$
    \textbf{Zongyuan Ge}\textsuperscript{1}$^\dagger$
\\[8pt]
        $^1$Monash University \quad
        $^2$Shanghai AI Laboratory \quad
        $^3$MBZUAI \\
        $^4$Imperial College London \quad 
        $^5$Eye Hospital, Wenzhou Medical Univeristy \\ 
        \texttt{ming.hu@monash.edu, z.yu23@imperial.ac.uk}\\
        \url{https://ophnet-3d.github.io/} \\
}

\begin{document}
\maketitle
\begin{abstract}
Accurate 3D reconstruction of hands and instruments is critical for vision-based analysis of ophthalmic microsurgery, yet progress has been hampered by the lack of realistic, large-scale datasets and reliable annotation tools. In this work, we introduce OphNet-3D, the first extensive RGB-D dynamic 3D reconstruction dataset for ophthalmic surgery, comprising 41 sequences from 40 surgeons and totaling 7.1 million frames, with fine-grained annotations of 12 surgical phases, 10 instrument categories, dense MANO hand meshes, and full 6-DoF instrument poses. To scalably produce high-fidelity labels, we design a multi-stage automatic annotation pipeline that integrates multi-view data observation, data-driven motion prior with cross-view geometric consistency and biomechanical constraints, along with a combination of collision-aware interaction constraints for instrument interactions. Building upon OphNet-3D, we establish two challenging benchmarks—bimanual hand pose estimation and hand–instrument interaction reconstruction—and propose two dedicated architectures: H-Net for dual-hand mesh recovery and OH-Net for joint reconstruction of two-hand–two-instrument interactions. These models leverage a novel spatial reasoning module with weak-perspective camera modeling and collision-aware center-based representation. Both architectures outperform existing methods by substantial margins, achieving improvements of over 2mm in Mean Per Joint Position Error (MPJPE) and up to 23\% in ADD-S metrics for hand and instrument reconstruction, respectively.
\end{abstract}
    
\section{Introduction}
\label{sec:intro}
Modern ophthalmic microsurgery represents one of the most delicate surgical paradigms in medicine, requiring sub-millimeter precision in instrument manipulation under restricted workspace conditions~\cite{pitcher2012robotic, tamai1993history}. While advances in robotic tools and surgical training platforms have improved treatment outcomes, current skill assessment methods still rely heavily on expert supervision and subjective feedback, limiting their scalability and objectivity~\cite{entezami2012mentorship, kotsis2013application, mao2021immersive}. In current ophthalmic surgical training paradigms, trainees predominantly rely on direct supervision from instructors for skill acquisition and performance evaluation. However, this approach imposes significant demands on instructional resources, particularly considering the time-intensive nature of surgical mentorship and the critical requirement for real-time feedback in complex microsurgical procedures. Recent studies in computer-assisted surgery~\cite{khalid2020evaluation, sugiyama2018forces, fujii2022surgical, goodman2024analyzing} reveal that kinematic analysis of surgical tools and operator hand movements could enable objective skill evaluation, personalized training feedback, and even real-time intraoperative guidance~\cite{sugiyama2018forces, WANG2025103599}. Goodman et al.~\cite{goodman2024analyzing} curated the 1,997-video AVOS corpus and trained a real-time multitask model that parses hands, tools, and actions. Building on AVOS, Vaid et al.~\cite{pmlr-v219-vaid23a} reframed surgeon-hand recognition as a semi-supervised, single-class 2D detection task that mixes many noisy unlabeled frames with a few labeled ones. Both efforts still rely on ~10-frame 2D snippets and lack long-range temporal, depth, pose, or multi-view cues. Meanwhile, other approaches~\cite{li2024extended, ozsoy2025mm} continue to depend on external motion-capture rigs or wearable sensors, introducing constraints that conflict with sterile surgical environments and disrupt natural workflow.

Recent advances in monocular~\cite{dong2024hamba, zhang2021interacting, wu2024reconstructing, pang2024sparse} and multi-view~\cite{yang2024mlphand, yang2023poem} RGB-based 3D hand-object interaction reconstruction have demonstrated notable success in general-purpose scenarios, offering a promising avenue for contactless skill assessment in ophthalmic surgical training. Estimating surgeons' hand and instrument poses from a single RGB image enables the quantification of critical operational details—such as grip posture and tool orientation—that are closely linked to surgical quality and clinical outcomes. However, directly applying these methods to ophthalmic microsurgical settings remains challenging due to the highly constrained operating space, fine-grained motion scale, and frequent occlusions and complex interactions between both hands and multiple instruments. These factors pose significant difficulties for accurate motion structure reconstruction using existing algorithms. Moreover, the lack of high-precision, realistically annotated 3D datasets specific to ophthalmic surgery further limits methodological development and evaluation in this domain.

To address the aforementioned limitations, we introduce OphNet-3D, the first large-scale dataset capturing dynamic 3D hand–instrument interactions in real-world ophthalmic surgeries. Collected with eight synchronized RGB-D cameras, it comprises 41 cataract surgery sequences by 40 surgeons (avg. >12 min/sequence), totaling over 7.1 M aligned RGB–D frames annotated for 12 surgical phases and 10 instrument categories. We apply a multi-stage automatic annotation pipeline to recover dense 3D hand meshes and 6D tool poses from these multi-view videos. Finally, we define two evaluation benchmarks—bimanual hand-pose estimation and hand–instrument interaction—and propose a unified baseline, OH-Net, for joint reconstruction of two-hand–two-tool interactions. Our contributions are:
\vspace{-2mm}
\begin{itemize}[leftmargin=*]
\item[$\bullet$]  
We present OphNet-3D, the first large-scale, real-world, high-quality dataset for 3D reconstruction of hand–instrument interactions in clinical surgical settings. OphNet-3D delivers an unprecedented combination of dataset scale, camera views, participant diversity, instrument, motion and object variety, and supported task types, at \textbf{2.5$\times$} the size of the largest existing general 3D hand reconstruction dataset and \textbf{70$\times$} that of the largest surgical 3D hand reconstruction dataset.

\item[$\bullet$] We propose a multi-stage automatic annotation pipeline that reconstructs 3D hand meshes and 6D tool poses from multi-view RGB-D videos using optimization with data-driven hand motion prior combined with biomechanical constraints and interaction-aware refinement.

\item[$\bullet$] Based on the proposed dataset, we establish two benchmarks: one for bimanual hand pose estimation and another for hand-tool interaction. We further introduce a unified baseline, OH-Net, which jointly reconstructs two-hand–two-tool interactions with effective spatial reasoning, and demonstrate its performance through extensive quantitative and qualitative results.


\end{itemize}

\vspace{-2mm}
\section{OphNet-3D Dataset}
\label{sec:ophnet3d}
\vspace{-2mm}
\begin{table*}[t!]
    \centering
    \caption{\textbf{Comparison with existing 3D hand reconstruction datasets.} OphNet-3D is the first surgical RGB-D dataset, offering high-resolution videos and rich annotations of complex hand–instrument interactions. It far exceeds prior datasets in scale and diversity and uniquely supports real-time dual-hand and multi-object reconstruction tasks. Task abbreviations: HPE: Hand Pose Estimation, OPE: Object Pose Estimation, HR: Hand Reconstruction, HOI: Hand-Object Interaction Reconstruction, HMOI: Hand and Multi-Object Interaction Reconstruction, Video: Video-level Reconstruction.}
\resizebox{\textwidth}{!}{
    \begin{tabular}{l|cccccccc|cccccc}
    \toprule
    & \multicolumn{8}{c|}{\large \textbf{Dataset Properties}} 
    & \multicolumn{6}{c}{\large \textbf{Task Support}} \\
     \textbf{Datasets}&  \textbf{Modality}&  \textbf{Source}&  \textbf{Views} & \textbf{Resolution}  & \textbf{Participants} &  \textbf{Obejects} &  \textbf{Motions} &  \textbf{Frames}  &  \textbf{HPE} & \textbf{OPE} &  \textbf{HR} &\textbf{HOI} &\textbf{HMOI} & \textbf{Video}\\
    \midrule
    FreiHAND~\textcolor{gray}{\scriptsize[ICCV'19]}~\cite{zimmermann2019freihand} & General & Real RGB & 8 &224$\times$224  & 32 & - & - & 130.2K & \ding{51} & \ding{55} & \ding{51} & \ding{55} & \ding{55} & \ding{51} \\
    ObMan~\textcolor{gray}{\scriptsize[CVPR'19]}~\cite{hasson19_obman} & General & Real RGB & 1 & 256$\times$256 & - & 8 & - & 150K & \ding{51} & \ding{51} & \ding{51} & \ding{51} & \ding{55} & \ding{55}\\
    InterHand2.6M~\textcolor{gray}{\scriptsize[ECCV'20]}~\cite{moon2020interhand2} & General & Real RGB & 80-140 & 512$\times$334 & 26 & 32 & - & 2.6M & \ding{51} & \ding{55} & \ding{51} & \ding{55} & \ding{55} & \ding{51}\\
    ContactPose~\textcolor{gray}{\scriptsize[ECCV'20]}~\cite{brahmbhatt2020contactpose}& General & Real RGB-D & 3  & 256$\times$256 & 50 & 25 & 2 & 2.9M & \ding{51} & \ding{51} & \ding{51} & \ding{51} & \ding{55} & \ding{51}\\
    H2O~\textcolor{gray}{\scriptsize[ICCV'21]}~\cite{Kwon_2021_ICCV} & General & Real RGB-D & 5 & 1280$\times$720 & 4 & 8 & 36 &571.6K& \ding{51} & \ding{51} &  \ding{51} & \ding{51} &\ding{55} & \ding{51} \\
    DexYCB~\textcolor{gray}{\scriptsize[CVPR'21]}~\cite{chao2021dexycb} & General &Real RGB-D & 8 & 640$\times$480 & 10 &20 & - & 582K & \ding{51} & \ding{51} &  \ding{51} & \ding{51} &\ding{55} & \ding{51} \\
    ARCTIC~\textcolor{gray}{\scriptsize[CVPR'23]}~\cite{fan2023arctic} & General &Real RGB& 9 & 2800$\times$2000 & 10 & 11 & 2 & 2.1M & \ding{51} & \ding{51} &  \ding{51} & \ding{51} &\ding{55} & \ding{51}\\
    HOT3D~\textcolor{gray}{\scriptsize[CVPR'25]}~\cite{banerjee2024hot3d} & General & Real RGB(mocap)& 3 & 1408$\times$1408 & 19 & 33 & - & 1.5M & \ding{51} & \ding{51} & \ding{51} & \ding{51} & \ding{55} & \ding{51} \\
    Hein et al.~\textcolor{gray}{\scriptsize[IJCARS'21]}~\cite{hein2021towards} & Clinical & Synth RGB & 2 & 256$\times$256 & 2 & 1 & - & 10.5K & \ding{51} & \ding{51} & \ding{51} & \ding{51} & \ding{55} & \ding{55}\\
    POV-Surgery~\textcolor{gray}{\scriptsize[MICCAI'23]}~\cite{wang2023pov} & Clinical & Synth RGB-D & 3 & 1920$\times$1080 & - & 3 & 3 & 88.3K &  \ding{51} & \ding{51} & \ding{51} & \ding{51} & \ding{55} & \ding{51}   \\
    HUP-3D~\textcolor{gray}{\scriptsize[MICCAI'24]}~\cite{birlo2024hup} & Clinical  & Synth RGB & 90 & 848$\times$480 & - & 1 & 11 & 31.7K  &  \ding{51} & \ding{51} & \ding{51} & \ding{51} & \ding{55} & \ding{55} \\
    \midrule
    OphNet-3D (Ours) & Clinical & Real RGB-D & 8 & 848$\times$480 & 40 & 10 & 12 & 7.1M & \ding{51} & \ding{51} & \ding{51} & \ding{51} & \ding{51} & \ding{51} \\
    \bottomrule
    \end{tabular} 
    }

\vspace{-0.4cm}
\label{tab:comp_general}
\end{table*}

\begin{figure}[t!]
  \centering
  \begin{tabular}{%
      @{} 
      p{0.63\textwidth}    
      @{\hspace{0.5em}\VertDashedLine{5.1cm}\hspace{0.2em}}
      p{0.36\textwidth}    
      @{}%
    }
    \vspace{0pt}
    \includegraphics[width=\linewidth]{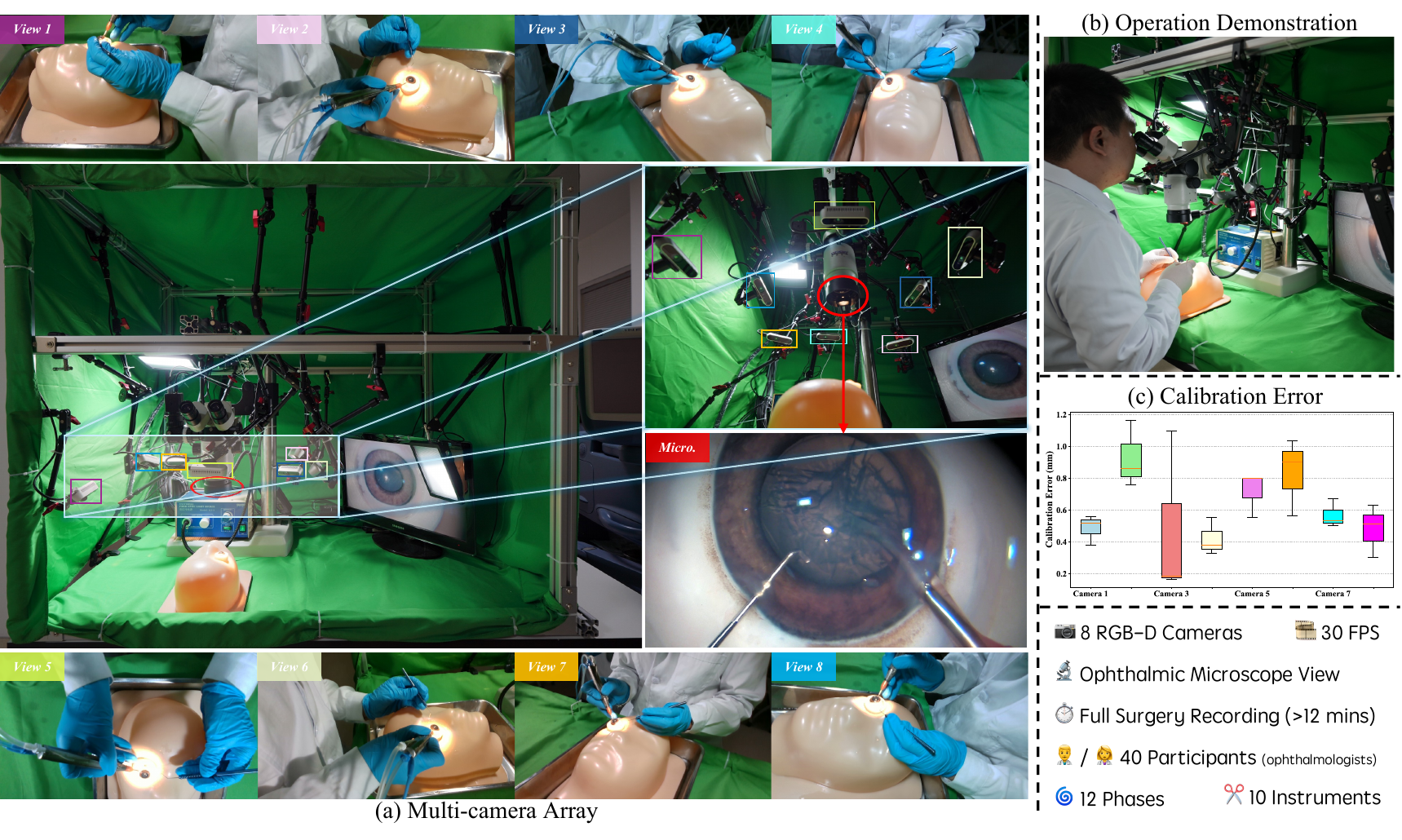}
    &
    \begin{minipage}[t]{\linewidth}
      \vspace{0pt}
      \includegraphics[width=\linewidth]{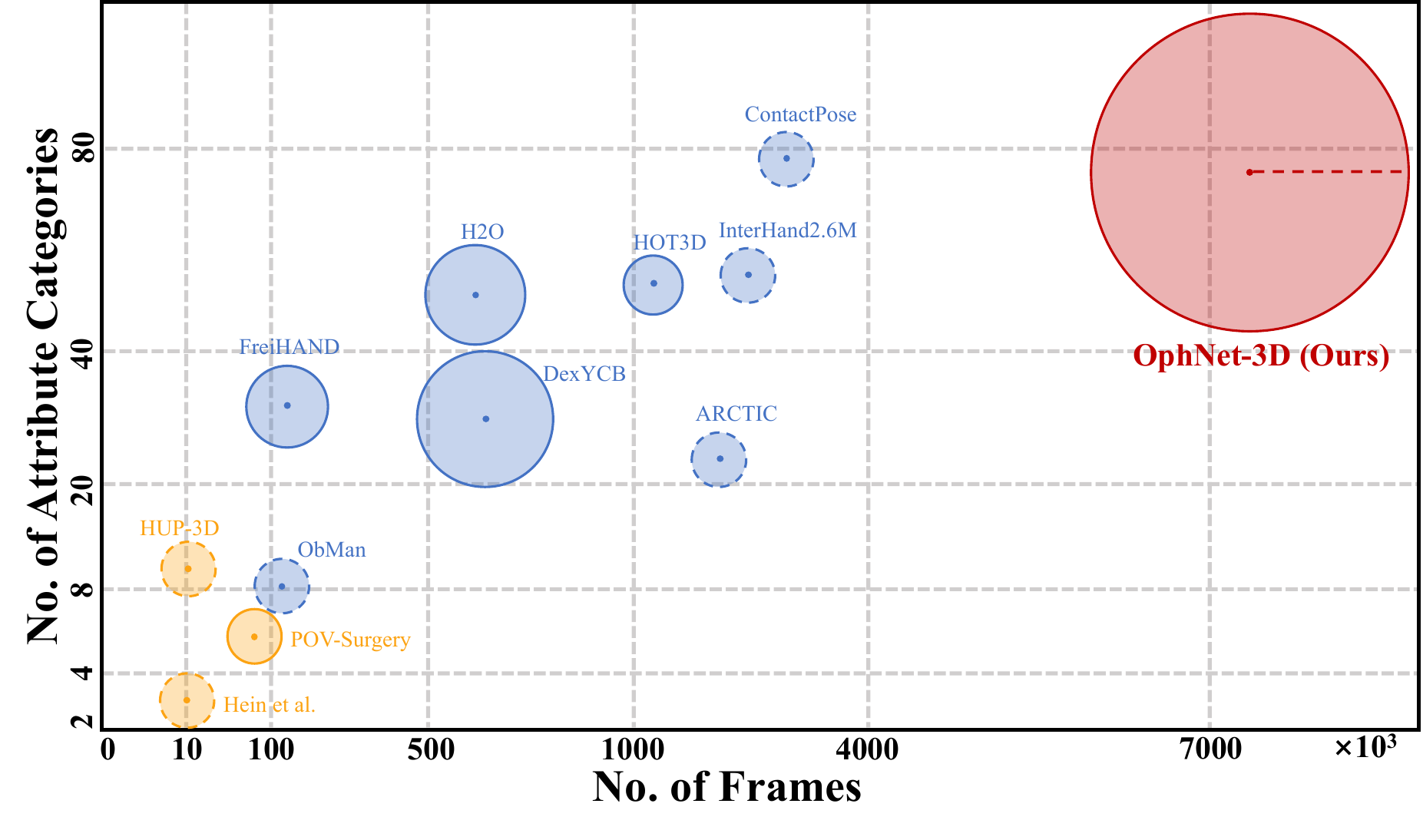}\\[-1.5mm]
      \includegraphics[width=\linewidth]{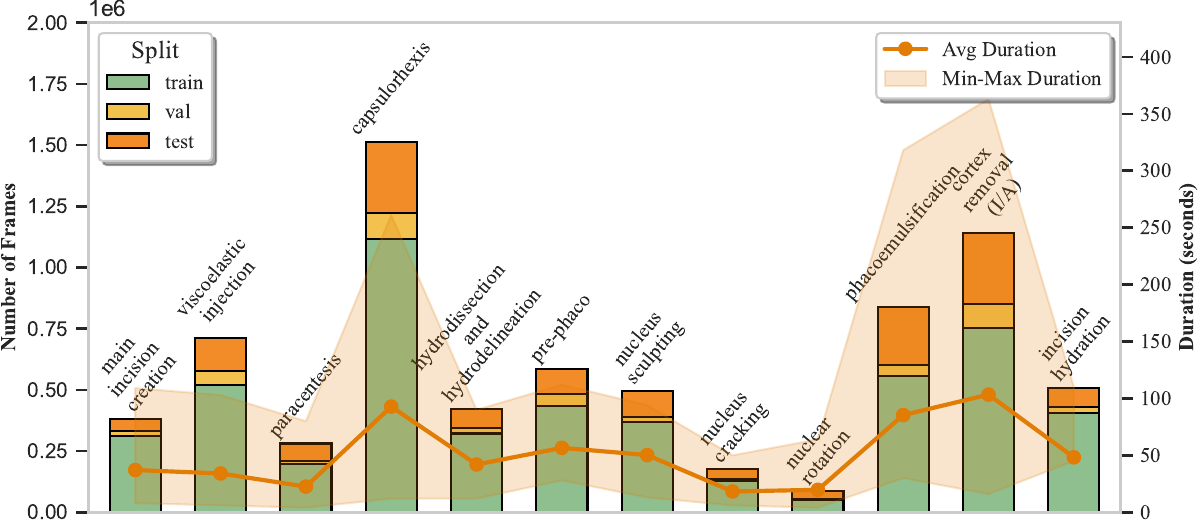}
    \end{minipage}
  \end{tabular}
  \caption{\textbf{OphNet-3D’s acquisition framework, comparisons with other datasets, and phase–frame distributions.} \textbf{\emph{Left:}} (a) a synchronized multi-camera rig with 8 calibrated RGB-D cameras and 3 directional LED lights; (b) participants perform standardized cataract surgery maneuvers on pig-eye simulators under an ophthalmic microscope; (c) boxplots of pixel errors for eight cameras across three calibration runs. \textbf{\emph{Upper Right:}} comparative visualization of OphNet-3D and existing 3D hand datasets. The horizontal axis denotes the total number of RGB frames, while the vertical axis indicates the number of categories across the participants, objects, and motion settings. Circle diameters encode the number of provided segmentation-mask instances; datasets without mask annotations are represented by dashed outlines. \textbf{\emph{Lower Right:}} distribution of frame counts and clip durations for each phase. }


  \vspace{-0.6cm}
  \label{fig:platform}
\end{figure}

\noindent \textbf{Data Collection.} 
\label{sec:ophnet3d_collection}
OphNet-3D is captured in a multi-camera studio consisting of 8 Intel® RealSense™ D435 RGB-D cameras recording at 30 FPS, along with 3 high-powered directional LED lights aimed at the hands to ensure uniform illumination (Fig.~\ref{fig:platform}: left). The cameras capture at a resolution of 848×480, and the multi-view system is calibrated using an ArUco calibration board. The detailed setup of the recording platform can be found in~\ref{configuration}.


We adhered to standard cataract surgical protocols, segmenting the procedure into 12 distinct phases. Detailed definitions and demonstration for each phase are provided in~\ref{phase_definition}. During the procedures, 10 different surgical instruments were employed and all instruments were scanned using a ZEISS ATOS Q blue-light 3D scanner, with corresponding images of the physical instruments and their 3D CAD models presented in~\ref{instrument}. During each recording session, two additional assistants were present—one operated the recording system, while the other assisted with the surgical workflow, such as instrument handover, to ensure procedural continuity. All surgical actions were recorded on an ophthalmic surgical simulation platform utilizing pig eyes, accompanied by synchronized video captured from an ophthalmic surgical microscope perspective. Finally, all videos were temporally aligned between the hand-view and microscope-view by an ophthalmologist, who also annotated the surgical phase locations and performed a secondary verification.


\noindent \textbf{Data Statistics.} 
We recorded a total of 41 sequences from 40 unique participants (one participant contributed two sequences, wearing blue and white gloves respectively), of whom 20 have more than one year of surgical experience and 20 have less than one year. Raw videos in our dataset have an average duration of ~16 minutes, comprising over 9.5M RGB frames. After filtering out transitional segments via phase localization annotation (~\ref{phase_localization}), the final OphNet-3D contains 565 phase segments with a total duration of 300 hours and over 7.1M RGB frames, as detailed in Tab.~\ref{tab:data_split_supp}. In addition, OphNet-3D provides segmentation annotations for more than 21M instances.


\vspace{-2mm}\section{Automatic Annotation Method} 
\begin{figure}[t]
    \centering
    \includegraphics[width=\textwidth]{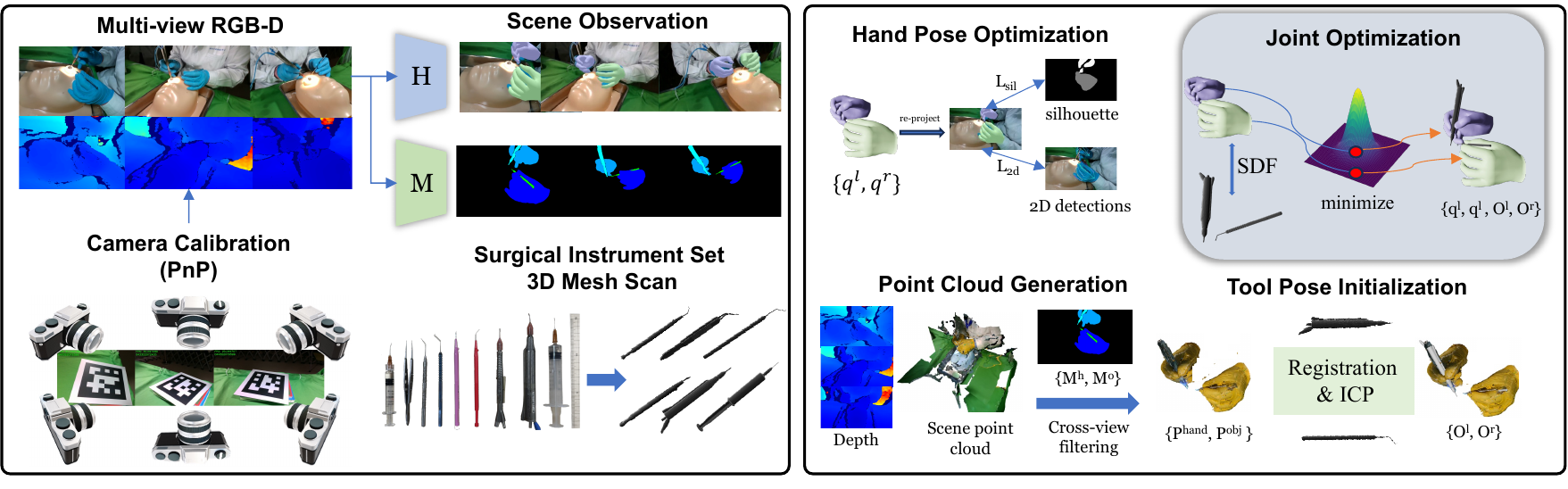}
    \caption{\textbf{Our automatic annotation pipeline.} Given a multi-view RGB-D video sequence as input, our pipeline reconstructs the 3D hand mesh and 6D instrument pose in a multi-stage manner. \textbf{H} and \textbf{W} represent the initialization network for hands \cite{pavlakos2024reconstructing} and instance segmentation masks \cite{ravi2024sam2}.}
    \label{fig:fitting} 
    \vspace{-0.5cm}
\end{figure}

\vspace{-2mm}
Given the input videos $\{\vid^{i}\}^{I}_{i=1}$ from multiple views with $T$ frames containing two hands interacting with possibly two surgical instruments, we aim to reconstruct the 3D hand-instrument interacting motions by recovering the hand meshes and 6D instrument poses. 
To efficiently auto-label the captured RGB-D videos with accurate mesh and pose annotations from multiple views, we design an optimization-based multi-stage automatic annotation pipeline as shown in \cref{fig:fitting}. 

In the first stage, we use the 3D CAD models scanned as described in Sec.~\ref{sec:ophnet3d_collection} to track per-frame 6D instrument poses. Moreover, we leverage the state-of-the-art SAM2 \cite{ravi2024sam2} with manual corrections to obtain per-frame accurate instance masks for hands ($^{i}\mathcal{M}_{h}$) and instruments ($^{i}\mathcal{M}_{o}$) for further point cloud segmentation. For per-frame scene point cloud generation, we first compute a point cloud from depth images for each view and merge the point clouds across views. To segment out the region of interest (\ie two hands and instruments) with cross-view filtering, we project the merged point cloud back to the 8 views and keep the points that project onto the hand-instrument region for more than half of the views to get the final per-frame scene point cloud $\mathbf{P}_{t}$, which can be further split into hand $\mathbf{P}^{hand}_{t}$ and instrument $\mathbf{P}^{obj}_{t}$.
In the second stage, our goal is to reconstruct the 3D hand and 6D instrument pose from the multi-view RGB-D videos recorded by 8 calibrated cameras. To this end, we leverage the state-of-the-art 2D \& 3D hand pose estimation method \cite{pavlakos2024reconstructing,lugaresi2019mediapipe} to initialize per-frame hand motion state in the camera coordinate system, as well as utilizing the masks for global registration and ICP \cite{lepetit2009ep} to estimate an initial instrument pose for each camera view.
Recovering accurate hand-instrument interaction is challenging due to frequent occlusions, truncation and mutual confusion. As a remedy, we propose a hand-instrument joint optimization scheme with a hand motion prior model HMP \cite{duran2024hmp} and biomechanical constraints in the third stage inspired by \cite{yu2024dyn}.





\vspace{-2mm}
\subsection{Hand Motion Annotation}
\vspace{-2mm}
\paragraph{Hand Representation.} 
We parametrize the hand shape and pose using the MANO hand model~\cite{MANO:SIGGRAPHASIA:2017},
which uses standard vertex-based linear blend skinning with learned blend shapes. At each time step $t$, the hand motion state is represented as:
\begin{equation}
\hand^h_t = \{\pose^h_t, \shape^h_t, \orient^h_t, \trans_t^h \},
\end{equation}
where $\pose^h_t \in \mathbb{R}^{3 \times 15}$ denotes the local pose of 15 hand joints, $\shape^h_t \in \mathbb{R}^{10}$ represents the hand shape parameters, and $(\orient^h_t, \trans_t^h)$ define the global wrist state. Specifically, the orientation $\orient^h_t \in \mathbb{R}^3$ is expressed using the axis-angle representation, while the translation $\trans_t^h \in \mathbb{R}^3$ specifies the wrist position in 3D space. The handedness is indicated by $h \in \{l, r\}$, representing left or right hand, respectively. Using these MANO parameters and the skinning function $\mathcal{W}(\cdot)$, we can reconstruct the 3D hand mesh $\handvert^h_t \in \mathbb{R}^{3 \times 778}$ and the 3D hand keypoints $\handjoints^h_t \in \mathbb{R}^{3 \times 21}$ with $\handvert^h_t = \mathcal{W}(\mathcal{H}(\joints^{h}_{t}, \shape^{h}), \mathcal{P}(\shape^{h}), \skinweight) + \trans^{h}_{t} \one_{778}$ and $\handjoints_{t}^{h} = \Lmat \handvert^h_t$.
where $\mathcal{W}(\cdot)$ denotes the skinning function, $\mathcal{H}$ represents the posed parametric hand template, and ${\one}_{778} \in \mathbb{R}^{1 \times 778}$ is a row vector of ones. The function $\mathcal{P}(\cdot)$ returns the hand joint positions in the rest pose, $\skinweight$ defines the skinning weights, and $\Lmat$ is a pre-trained linear regressor for estimating joint locations from mesh vertices. 

\paragraph{Hand Initialization.} For each camera view, we initialize per-frame 3D hand motion state in camera coordinate system leveraging an efficient two-hand motion tracking system based on \cite{pavlakos2024reconstructing} with hallucinated detection handling. 
We further obtain per-frame 2D hand keypoints from ViTPose \cite{xu2022vitpose} and hand bounding box to extract image patches, feeding into \cite{pavlakos2024reconstructing} for a coarse-to-fine 3D motion state prediction. Finally, we compute the weighted sum of the motion state based on the visibility to obtain the final initilization of the global motion state $^{i}\hand^{h}_{t}$ in the world coordinate system.
Next, we convert the motion state into the world coordinate system using the calibrated camera information $\{\Rot^{i}_{t}, \pos^{i}_{t}\}$ of camera view $i$: 
\begin{align}
    \worient^h_{t} = \Rot^{-1}_{t}{\cdot}\corient^{h}_{t} \qquad\mathrm{and}\qquad
    \wtrans^{h}_t = \Rot^{-1}_{t}  {\cdot}\ctrans^{h}_{t} -  \Rot^{-1}_{t}{\cdot}\pos_{t},
\end{align}
where $\worient^h_{t}$ and $\corient^{h}_{t}$ are the hand wrist orientation in world and camera space. $\wtrans^{h}_t$ and $\ctrans^{h}_{t}$ are the translation. Here we omit $i$ for simplicity.
For 2D observations, we initialize from ViTPose \cite{xu2022vitpose}, MediaPipe \cite{lugaresi2019mediapipe} and the 2D re-projection from \cite{pavlakos2024reconstructing} with a confidence filter to extract final per-frame 2D keypoint $^{i}\mathbf{\hat{\handjoints^{h}_{t}}} \in \mathbb{R}^{2\times 21}$ for each view as observation for following optimization, where the 2D re-projection is performed with weak-perspective camera parameters predicted by \cite{pavlakos2024reconstructing}. Moreover, we provide more details regarding the post-processing of the observations, including hallucination handling and missing detection infilling in the \ref{processing:appendix}.

\paragraph{Optimization.} To recover the hand meshes from multi-view RGB-D videos, we propose an iterative fitting algorithm by minimizing the following objective with regularization and \emph{biomechanical hand constraints}~\cite{spurr2020weakly}, as well as hand motion prior \cite{duran2024hmp}.
\begin{align}\label{eq:Eone}
    E_{\mathrm{I}}(\pose^h_t, \shape^h_t, \orient^h_t, \trans_t^h) = & \, \sum^{N_{i}}_{i=1}({\lambda_{\mathrm{2d}}\LTwoD + \lambda_{\mathrm{sil}}\Lsil}) + \lambda_{\mathrm{s}}\Lsmooth  + \lambda_{\mathrm{3d}}\LThreeD + \Lbio + \Lprior, 
\end{align}
where $N_{i}$ is the number of camera views and $\LTwoD$ is the joint 2D re-projection loss minimizing the difference between 2D hand keypoints observations $\{\hat{\handjoints^{h}_{t}}\}^{T}_{t=0}$ and the re-projection of the 3D keypoints obtained from MANO model with parameters $\{\pose^h_t, \shape^h_t, \orient^h_t, \trans_t^h \}$ in current global state $\hand^{h}_{t}$:
\begin{equation}\label{eq:L2d}
\centering
    \LTwoD = \sum_{h \in \{l, r\}}\sum_{t=0}^{T} \rho\left(\conf^{h}_{t}\left(\Tilde{^{i}\handjoints_{t}^{h}}-\hat{^{i}\handjoints_{t}^{h}}\right)\right).
\end{equation}
where $\rho(\cdot)$ is the Geman-McClure robust function \cite{geman1987statistical}. $\conf^{h}_{t}$ is a confidence filter mask for joint visibility. $\Tilde{\handjoints_{t}^{h}} = \Pi(\whandjoints_{t}^{h}, \Rot_{t}, \pos_{t}, \K)$ is the re-projected 3D keypoints, and $\Pi$ is the perspective camera projection for each view with collected camera intrinsics $\K \in \mathbb{R}^{3 \times3}$ and extrinsics $\{\Rot_{t}^{i}, \pos_{t}^{i}\}$. 
To improve the pixel alignment across views, we propose a silhouette-based error term $\Lsil$:
\begin{equation}\label{eq:Lsil}
\centering
    \Lsil = \sum_{t=0}^{T}\|\ ^{i}\mathbf{M}_{t}^{h} - \mathbf{NR}(V_{t}^{h}, K^{i}, \Rot_{t}^{i}, \pos_{t}^{i}) \|
\end{equation}
where $\mathbf{NR}(\cdot)$ is a differentiable renderer that renders the 3D mesh for the two hands into the 2D mask and $\mathbf{M}_{t}^{h}$ is the clean segmentation mask for the hand $h$ at timestep $t$. To have a precise 3D alignment between the mesh reconstruction and the real scene, we compute the 3D mesh loss and minimize the distance between the frame point cloud $\mathbf{P}_{t}^{h}$ and the MANO hand mesh:
\begin{equation}
    \LThreeD = \sum_{h \in \{l, r\}} \sum_{t=0}^{T} \sum_{i=1}^{N_V} \frac{1}{N_V} \|\left( \mathbf{p}^{j(i)}_{t} - \mathbf{v}^i_t \right) \cdot \mathbf{n}_i \|,
\quad \text{where} \quad j(i) = \arg\min_{j} \| \mathbf{p}^{j}_{t} - \mathbf{v}^{i}_{t} \|
\end{equation}
where $\mathbf{p}^{j}_{t}$ is the $j$-th point of the hand point cloud $\mathbf{P}_{t}^{h}$ at timestep $t$ and $v^{i}_{t}$ is the $i$-th vertex of the hand mesh $\mathbf{V}_{t}^{h}$. The reconstructed hand mesh normal of the vertex is represented as $n_{t}^{i}$. Moreover, we reduce the jitter of the hand motion and improve the temporal smoothness by integrating $\Lsmooth$:
\begin{equation}
    \Lsmooth = \sum_{h \in \{l, r\}}\sum_{t=0}^{T}\|\handjoints_{t+1}^{h} - \handjoints_{t}^{h}\|^{2} + g(\pose_{t+1}^{h}, \pose_{t}^{h})^{2}
\end{equation}
where $g(\cdot)$ represents the geodesic distance. To further improve the plausibility of hand motion qualityand reduce jitter for natural movement, we compute the $\Lprior=\mathcal{L}_{\z} + \lambda_{\theta}\Lpose + \lambda_{\beta}\Lshape$ by utilizing a data-driven motion prior \cite{duran2024hmp} where the latent code is $\z^{h}$ Inspired by \cite{yu2024dyn}. It ensures the motion is constrained under the learnt prior space, by penalizing the negative log-likelihood:
\begin{equation}
    \mathcal{L}_{\z} = \sum_{h\in \{l ,r\}}\sum_{t=0}^{T} - \log\mathcal{N}(\\ \z^{h};\mu^{h}(\{\handjoints^{h}_{t}\}), \sigma^{h}(\{\handjoints^{h}_{t}\})).\nonumber
\end{equation}
To explicitly prevent implausible poses produced during optimization, we further propose $\Lbio = \sum_{t=0}^{T}(\lambda_\mathrm{ja}\Lja + \lambda_\mathrm{bl}\Lbl + \lambda_\mathrm{palm}\Lpalm + \lambda_\mathrm{angle}\Langle$), which consists of angle regularization terms with biomechanical constraints~\cite{spurr2020weakly} and and an angle limitation constraint. More details regarding the loss calculation are provided in \ref{pipeline:appendix}.

\vspace{-1mm}\subsection{Instrument Motion Annotation}\vspace{-1mm}\label{sec:instrument_annotation}
By leveraging multi-view RGB-D frames together with camera pose information, our method can accurately annotate per-frame 6D instrument pose of the instruments. To recover 3D hand motion under the challenging surgical scenario (\eg light, occlusion).

\paragraph{Obtaining Canonical Local Instrument Geometry.}  
We laser-scanned each instrument to obtain high-resolution 3D meshes (Sec.~\ref{sec:ophnet3d}). For articulated instruments (\eg \emph{phacoemulsification handpiece}), we additionally separate and scan them into two articulated parts, as well as their rest pose and maximum relative pose articulation as shown in \cref{fig:fitting}. Please see~\ref{fig:instrument_3d} for the detailed visualization of the instruments.



\paragraph{Acquiring Instrument Articulation.}
The articulated instrument (\eg handpiece) surface is parameterised by the 6D pose of each base part and a 1D articulation relative to a canonical pose. In particular, the 6D instrument pose can be represented as $\{\Rot^{o}_{t}, \pos^{o}_{t}\}$. For each instrument, we define a 3D parametric model $\mathcal{O}(\cdot)$ leveraging the scanned instrument parts and relative pose state. Given the 6D pose $\theta^{o}_{t} \in \mathbb{R}^6$ and the 1D relative articulation factor $\alpha_{t} \in \mathbb{R}^1$, where $\alpha \in [0, 1]$ uniformly parameterizes pose deformation across different models. Here, $\alpha = 0$ corresponds to the rest pose, and $\alpha = 1$ to the maximally articulated pose, representing the deformation state. The instrument 3D mesh $\mathcal{O}(\theta^{o}_{t},\alpha_{t}) \in \mathbb{R}^{3 \times N_{o}}$ can be reconstructed, where $N_{o}$ is the instrument vertices number. 

\paragraph{Initialization of 6D instrument pose.} We obtain accurate per-frame 6D instrument pose leveraging the multi-view RGB-D information. As mentioned, we first perform instrument segmentation and 2D tracking for each camera view using SAM2 \cite{ravi2024sam2} with manual correction to obtain clean 2D segmentation masks for both hands and instruments. Moreover, we merge the depth image across views to generate the point cloud for the whole scene. After that, we segment out the region of interest leveraging a cross-view filter, which projects the point cloud into all camera views and keeps the points projected onto the hand-instrument region for more than half of the views to get the point cloud for instruments $\mathbf{P}^{obj}_{t}$. By running RANSAC-based global registration, we have a coarse global alignment of the 3D instrument mesh and $\mathbf{P}^{obj}_{t}$. After that, we refine the alignment with ICP \cite{lepetit2009ep} to obtain the initial rigid transformation from instrument canonical coordinate system to the world coordinate system. Finally, we run a simple Chamfer distance-based optimizer for better alignment and to obtain the articulation $\alpha_{t}$ and the final 6D instrument pose $\{\Rot^{o}_{t}, \pos^{o}_{t}\}$. By applying the 6D pose to the instrument model, we can obtain the 3D mesh in the world coordinate system. Note that our surgical scenario contains various two-hand-two-instrument interactions, thus we represent the instrument in each hand as $\mathbf{O}^{h}_{t}$.

\vspace{-1mm}\subsection{Joint Optimization}\vspace{-1mm} Naively putting the hand and instrument together may result in implausible hand-instrument interactions such as inter-penetration and unnatural contact. To jointly optimize 3D hand pose and instrument pose and introduce more constraints for the interaction, we propose the following objectives:
\begin{align}\label{eq:Eobj}
    E_{\mathrm{II}}(\pose^h_t, \shape^h_t, \orient^h_t, \trans_t^h, \Rot^{o}_{t}, \pos^{o}_{t}, \alpha_{t}) = & \ E_{I} + 
 \sum^{N_{i}}_{i=1}({\lambda_{\mathrm{sil}}\Lsil}) + \lambda_\mathrm{3d}\LThreeD + \lambda_{\mathrm{inter}}\Linter  + \lambda_{\mathrm{sdf}}\Lsdf
\end{align}
Specifically, $\Lsil$ represents the silhouette loss term that is computed between the combined hand-object mask $^{i}\mathbf{M}^{h,o}_{t}$ and the rendered mask $\mathbf{NR}(V^{h}_{t}, O_{t}^{h}, K^{i}, \Rot_{t}^{i}, \pos_{t}^{i})$ of the 3D hand-object mesh. $\LThreeD$ is calculated between ground truth point cloud $\mathbf{P}_{t}$ and the predicted hand and object mesh. Moreover, we leverage the interaction loss $\Linter$ to constrain the hand-instrument contact following \cite{hasson2019learning} as $\Linter = \lambda_{R}\mathrm{L}_{R} + \lambda_{A}\mathrm{L}_{A}$, where $\mathrm{L}_{R}$ and $\mathrm{L}_{A}$ are the attraction loss and the repulsion loss, which penalize the interpenetration between hand and instruments, and minimize the distance between the interacting hand and instrument in the possible contact region, respectively. We provide more details regarding the calculation in the Appendix. Finally, we refine the interaction by applying a modified version of Signed Distance Field (SDF) loss $\Lsdf$, for which we provide more details and an ablation study in  \ref{pipeline:appendix}.



\vspace{-2mm}
\section{Baseline and Experiments}
\label{sec:experiments}

Our dataset can enable various downstream tasks for pose estimation and recognition. In this section, we introduce two benchmarks built upon our dataset. We first propose the evaluation protocols of each benchmark and provide a detailed analysis of our dataset. Furthermore, we present baseline methods corresponding to the benchmarks with comparison against the state-of-the-art methods to demonstrate the effectiveness of our approach. More implementation details are provided in~\ref{experiment:appendix}.


\vspace{-2mm}
\subsection{Evaluation Protocol} 
\noindent \textbf{Data Split.} To ensure each phase has balanced samples, we split our dataset into
\begin{wrapfigure}[]{r}{0.55\textwidth}
\centering
        \includegraphics[width=\linewidth]{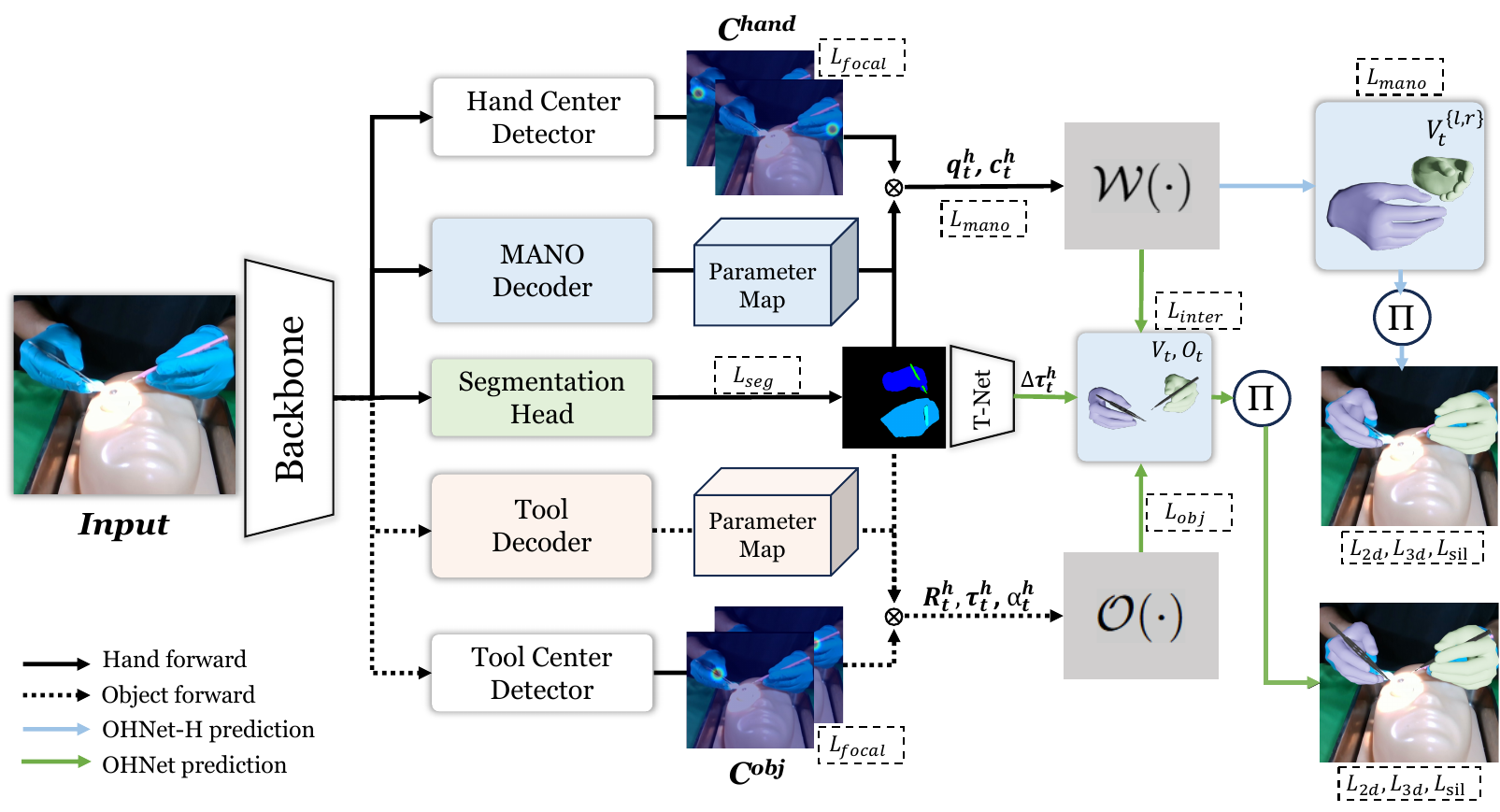}
        \vspace{-0.4cm}
        \caption{\textbf{Overview of the OH-Net.} The backbone image encoder outputs the image feature, which is then used to decode the hand/instrument centre heatmap and segmentation mask. MANO decoders predict their corresponding weak-perspective camera parameters. Decoupling the instrument branch forms \textbf{H-Net}.}
        \label{fig:baselines}
        \vspace{-0.6cm}
\end{wrapfigure}
training, validation, and test sets by subjects, which have 30, 3, 8 subjects separately. Based on the data split, bimanual hand pose estimation and (2) hand-instrument interactions. We provide more details regarding the data distribution and data quality analysis in the Appendix. Note that in our experiments on both benchmarks, we train the model on the monocular training images from all 8 views, including both egocentric and allocentric for rich supervision.

\noindent \textbf{Evaluation Metrics.} Our goal is to reconstruct accurate 3D motion of hands and instruments during complex surgical operations from video. Specifically, we propose metrics to quantify estimate quality and compare our baseline method against state-of-the-art hand-object pose approaches.


\begin{itemize}[leftmargin=*,noitemsep,topsep=0em]
    \item \textbf{Bimanual Hand Pose Estimation:} To evaluate the accuracy and plausibility of the hand reconstruction pipeline, we report the Mean Per Joint Positional Error (MPJPE), Mean Per Vertex Positional Error (MPVPE) in $mm$ after root (hand wrist) joint alignment. To explicitly measure the relative translation error, we report Mean Relative Root Translation Error (MRRTE). 
    \item \textbf{Hand-instrument Interaction:} To quantify the reconstruction quality, we report the same evaluation metrics (MPJPE, MPVPE, MRRTE) as in bimanual hand pose estimation benchmarks. For instrument pose estimation quality, we evaluate the commonly used ADD-S score, which measures the average distance between the model vertices transformed by the ground truth and the estimated poses, following \cite{lin2023harmonious,kwon2021h2o,stevvsivc2020spatial,tekin2019h+,zakharov2019dpod}. Specifically, we report the percentage of the transformed instruments with a vertex positional error less than 10\% of the instrument diameter. We further report the Mean Articulation Error (MAE) to evaluate the articulation of instruments, calculating the absolute error between the ground-truth articulation factor and the prediction in percentage, excluding the rigid instruments without articulation. For the interaction quality, we evaluate Mean Per Joint Positional Error between each hand-instrument interaction pair (MRRTE${h,o}$) and Mean Inter-penetration Volume (Pen) in $cm^{3}$.
\end{itemize}


\vspace{-2mm}\subsection{Bimanual Hand Motion Estimation}\vspace{-2mm}\label{sec:bimanual}
\begin{wraptable}{r}{0.55\columnwidth}  
  \vspace{-12pt}
  \centering
  \small   
  \renewcommand{\arraystretch}{1.1}
  \caption{\textbf{Quantitative evaluation results for bimanual hand motion estimation.} We compare our method with state-of-the-art hand reconstruction methods on local hand poses. MPJPE, MPVPE, and MRRTE are reported in millimeters (mm) after root alignment.}
  \label{tab:hand_val_test}
  \resizebox{\linewidth}{!}{%
    \begin{tabular}{l|ccc|ccc}
      \toprule
        \multirow{2}{*}{\textbf{Method}} & 
        \multicolumn{3}{c|}{\textbf{Val}} & 
        \multicolumn{3}{c}{\textbf{Test}} \\
      \cmidrule(lr){2-4}\cmidrule(lr){5-7}
       & MPJPE $\downarrow$ & MPVPE $\downarrow$ & MRRTE $\downarrow$
       & MPJPE $\downarrow$ & MPVPE $\downarrow$ & MRRTE $\downarrow$ \\
      \midrule
      DIR~\cite{ren2023decoupled}       & 18.63 & 18.91 & 34.76 & 18.89 & 18.75 & 35.17 \\
      InterWild~\cite{moon2023bringing} & 19.48 & 19.87 & 37.27 & 20.19 & 20.34 & 38.76 \\
      IntagHand~\cite{Li2022intaghand}  & 18.92 & \underline{17.96} & 32.43 & 19.38 & 19.16 & 32.77 \\
      ACR~\cite{yu2023acr}              & 18.18 & 18.57 & 33.29 & 18.86 & 19.28 & 33.59 \\
      \midrule
      \textbf{H-Net (w/o T-Net)}                & 18.57 & 18.48 & 33.92 & 18.98 & \underline{19.14} & 34.18 \\
      \textbf{H-Net}                     & \underline{17.39} & 18.72 & \underline{31.66} & \underline{17.66} & 18.66 & \underline{31.89} \\
      \textbf{H-Net-D}                  & \textbf{15.28} & \textbf{16.37} & \textbf{26.86} & \textbf{15.97} & \textbf{16.18} & \textbf{26.59} \\
      \bottomrule
    \end{tabular}
  }
  \vspace{-13pt}
\end{wraptable}
We now set up the benchmark for bimanual hand pose estimation from a monocular RGB input image. Acquiring accurate 3D hand pose is essential in the surgical scenarios during manipulation.

\textbf{Parametric Representation.} In the task of (monocular) bimanual hand pose estimation, our goal is to reconstruct the 3D pose of the two hands from an RGB input video. In order to obtain the detailed geometry of the two hands, we adopt the parametric model MANO \cite{MANO:SIGGRAPHASIA:2017} as our mesh representation to predict $\{\pose^h_t, \shape^h_t, \orient^h_t, \trans_t^h\}$ following the dataset settings. Given the MANO parameters, the 3D hand mesh $V^{h}_{t}$ and the 21 hand keypoints $\handjoints^{h}_{t}$ can be regressed. We adopt the commonly used weak-perspective camera model following \cite{pavlakos2024reconstructing,Li2022intaghand,yu2023acr,moon2020interhand2,zhang2021interacting} to estimate the 3D translation. 

\noindent\textbf{Baseline.} To address the problem of reconstructing bimanual hands from a monocular RGB(-D) image, we introduce H-Net as the baseline approach, without the instrument branch. As shown in \cref{fig:baselines}, the model takes as input the image at timestep $t$ to extract the image feature $f_{t} \in \mathbb{R}^{D \times H \times W}$ where $D$ is the dimension of the feature map. Subsequently, the following 3 regression heads predict the segmentation mask $\mathbf{M}^{h} \in \mathbb{R}^{3 \times \mathrm{H} \times \mathrm{W}}$ (\ie left hand, right hand, background), MANO parameter map $\mathbf{M}^{mano}_{t} \in \mathbb{R}^{218 \times \mathrm{H} \times \mathrm{W}}$, hand center heatmap $\mathbf{C}^{hand}_{t} \in \mathbb{R}^{2\times \mathrm{H} \times \mathrm{W}}$ respectively. Leveraging the collision aware center-based representation \cite{sun2021monocular,yu2023acr} for hands, we disentangle the bimanual hand features while pushing away the centers that are too close in the repulsion field. In the following, the MANO parameters $\hand_{t}^{h}$ for each hand is extracted by combining with the Hand Center map and the instance segmentation mask. After obtaining the 3D mesh by MANO model $\mathcal{W(\cdot)}$ with keypoints, we use the output relative translation $\Delta \tau \in \mathbb{R}^{3}$ from T-Net to model the fine-grained relative transformation from the left hand to the right hand, incorporating the strong spatial features as prior knowledge. The weak-perspective camera parameter is represented as $c_{t}^{h}$. Moreover, we denote the RGB-D input based version as H-Net-D. Finally, the network is supervised by the weighted sum of the hand center loss and the mesh parameter loss:
\begin{equation}
    \mathcal{L} = \lambda_\mathrm{focal}\mathcal{L}_{focal} + \lambda_\mathrm{pj2d}\LpjTwoD + \lambda_\mathrm{3d}\LThreeD + \lambda_\mathrm{sil}\Lsil + \Lmano + \lambda_\mathrm{seg}\Lseg
\end{equation}
where $\mathcal{L}_{focal}$ is the focal loss for the hand center map. $\Lmano=\lambda_{\theta} \Lpose + \lambda_{\beta} \Lshape$ is the weighted sum of L2 loss of the MANO parameters. We provide the training and implementation details in \cref{baselines:appendix}.


\paragraph{Results.} We evaluate the performance of our baseline models on surgical bimanual hand reconstruction tasks, and compare them against state-of-the-art hand pose estimation methods. As shown in \cref{tab:hand_val_test}, our method significantly outperforms prior approaches such as InterWild~\cite{moon2023bringing}, DIR~\cite{ren2023decoupled}, and IntagHand~\cite{Li2022intaghand}, achieving the best overall performance across different metrics, including MPJPE, MPVPE, and MRRTE. These results highlight the importance of domain-specific design: our hand center detector and tailored parameterization improve robustness in surgical environments, where factors such as gloves, occlusions, and instrument-induced hand articulation pose unique challenges. Notably, the T-Net module contributes to finer pose refinement by learning spatial alignment from segmentation masks, leading to consistent improvements across both joint-wise and vertex-level metrics. These findings validate the effectiveness of our baseline in modeling surgical hands with high precision, serving as a strong foundation for subsequent hand-instrument reasoning.

\vspace{-2mm}\subsection{Two-Hand-Instrument Interactions}\vspace{-1mm}
In this section, we propose the benchmark for hand-instrument interaction, which aims to reconstruct the 3D mesh for two hands and the 6D pose for the surgical instruments.

\paragraph{Parametric Representation} In terms of two-hand-instrument Interaction baseline, our task is to reconstruct the 3D meshes of the two hands as well as the 6D pose of the in-hand surgical instruments. We keep the hand representation as MANO \cite{MANO:SIGGRAPHASIA:2017} for consistency. For 6D instrument pose estimation, we leverage the parametric model $\mathcal{O}(\theta^{o}_{t}, \alpha_{t})$ introduced in \cref{sec:instrument_annotation} to represent the surgical instruments with articulation. Specifically, the 3D mesh is regressed using the parameters of 6D pose $\theta^{o}_{t} \in \mathbb{R}^6$ and the 1D relative articulation factor $\alpha_{t} \in [0, 1]$ which controls the articulation.

\begin{table}[t!]
\caption{\textbf{Quantitative evaluation results for two-hand-instrument interactions.} We compare our method with the state-of-the-art hand reconstruction methods on local hand poses.}
\label{tab:handobj}
\centering
\resizebox{\linewidth}{!}{
\begin{tabular}{ll|ccc|ccc|c}
\toprule
\textbf{Split} & \textbf{Method ($mm$)} & \textbf{MPJPE ($mm$)} $\downarrow$ & \textbf{MPVPE ($mm$)} $\downarrow$ & \textbf{MRRTE ($mm$)} $\downarrow$ & \textbf{ADD-S (\%)} $\uparrow$ & \textbf{MAE (\%)} $\downarrow$ & \textbf{Pen ($mm$)} $\downarrow$ & \textbf{MRRTE$^{h,o}$ ($mm$)} $\downarrow$ \\
\midrule
\multirow{5}{*}{Val} 
     & Hasson et al. \cite{hasson2020leveraging} & 19.87  & 21.45  & 39.78 & 56.66 & - & 7.87 & 27.98 \\
     & HFL-Net \cite{lin2023harmonious}         & 17.51  & 18.27  & 37.43 & 58.64 & - & 7.06 & 25.67 \\
     & HOISDF \cite{qi2024hoisdf}               & \underline{17.05} & \underline{18.22} & 34.36 & 60.91 & - & \underline{5.69} & 24.13 \\
     & \textbf{OH-Net (w/o T-Net)}              & 17.48 & 18.33 & 32.61 & 66.87 & 14.83 & 6.41 & 21.86 \\
     & \textbf{OH-Net}                          & \underline{17.12} & 18.43 & \underline{31.36} & \underline{71.52} & \underline{11.14} & 5.87 & \underline{19.94} \\
     & \textbf{OH-Net-D}                        & \textbf{15.23} & \textbf{16.41} & \textbf{26.78} & \textbf{76.68} & \textbf{9.67} & \textbf{5.14} & \textbf{17.33} \\
\midrule
\multirow{5}{*}{Test} 
     & Hasson et al. \cite{hasson2020leveraging} & 20.04 & 21.33 & 40.56 & 56.89 & - & 7.76 & 28.69 \\
     & HFL-Net \cite{lin2023harmonious}         & 17.45 & \underline{17.66} & 38.65 & 59.78 & - & 7.14 & 25.49 \\
     & HOISDF \cite{qi2024hoisdf}               & \underline{17.36} & 17.91 & 35.88 & 61.32 & - & \underline{5.77} & 23.82 \\
     & \textbf{OH-Net (w/o T-Net)}              & 17.89 & 18.06 & 33.25 & 65.94 & 14.31 & 6.45 & 21.92 \\
     & \textbf{OH-Net (Ours)}                   & \underline{17.34} & 18.36 & \underline{31.58} & \underline{70.79} & \underline{11.17} & 5.91 & \underline{20.11} \\
     & \textbf{OH-Net-D (Ours)}                 & \textbf{15.94} & \textbf{16.13} & \textbf{26.44} & \textbf{76.31} & \textbf{9.62} & \textbf{5.18} & \textbf{17.45} \\
\bottomrule
\end{tabular}

}
\vspace{-0.5cm}
\end{table}

\paragraph{Baseline.} As discussed in \cref{sec:bimanual}, integrating the instrument 6D pose estimation branch forms the full model of OH-Net. As the first method to reconstruct two-hand-two-object, we propose to disentangle the features and mesh representation along with an explicit handler for interaction. As illustrated in \cref{fig:baselines}, the extracted feature map is followed by 5 regression heads, yielding hand center map $\mathbf{C}^{hand}_{t} \in \mathbb{R}^{2\times \mathrm{H} \times \mathrm{W}}$ and object center map $\mathbf{C}^{obj}_{t} \in \mathbb{R}^{2\times \mathrm{H} \times \mathrm{W}}$. The segmentation head predicts the instance mask $\mathbf{M}^{h,o}_{t} \in \mathbb{R}^{5 \times \mathrm{H} \times \mathrm{W}}$. We extend the collision-aware center-based representation mechanism to fit in the task and push away the two close instrument center and hand centers. The instrument parameter map is regressed as $\mathbf{M}^{obj} \in \mathbb{R}^{14 \times \mathrm{H} \times \mathrm{W}}$, which contains the 6D pose and 1D articulation factor for both instruments. Next, T-Net predicts $\Delta \tau \in \mathbb{R}^{9}$ for the relative translation between the left hand, right hand, and between their interacting instruments. OH-Net is supervised with the objectives below:
\begin{equation}
    \mathcal{L} = \lambda_\mathrm{focal}\mathcal{L}_{focal} + \lambda_\mathrm{pj2d}\LpjTwoD + \lambda_\mathrm{3d}\LThreeD + \lambda_\mathrm{sil}\Lsil + \Lmano + \lambda_\mathrm{seg}\Lseg + \lambda_\mathrm{obj}\Ltool
\end{equation}
where $\Ltool$ represents the loss function for instrument supervision. Specifically, $\Ltool$ is composed of $\mathcal{L}_{R}$ and $\mathcal{L}_{\tau}$ for the 6D instrument pose and $\mathcal{L}_{\alpha}$ for the 1D articulation. Moreover, $\LpjTwoD$ and $\LThreeD$ is also defined for instrument and optimized with pre-defined bounding box of the instrument mesh following \cite{liu2021semi}. We refer the readers to the appendix for detailed implementation details.




\begin{wrapfigure}[18]{r}{0.45\textwidth}
\vspace{-4mm}
\centering
        \includegraphics[width=\linewidth]{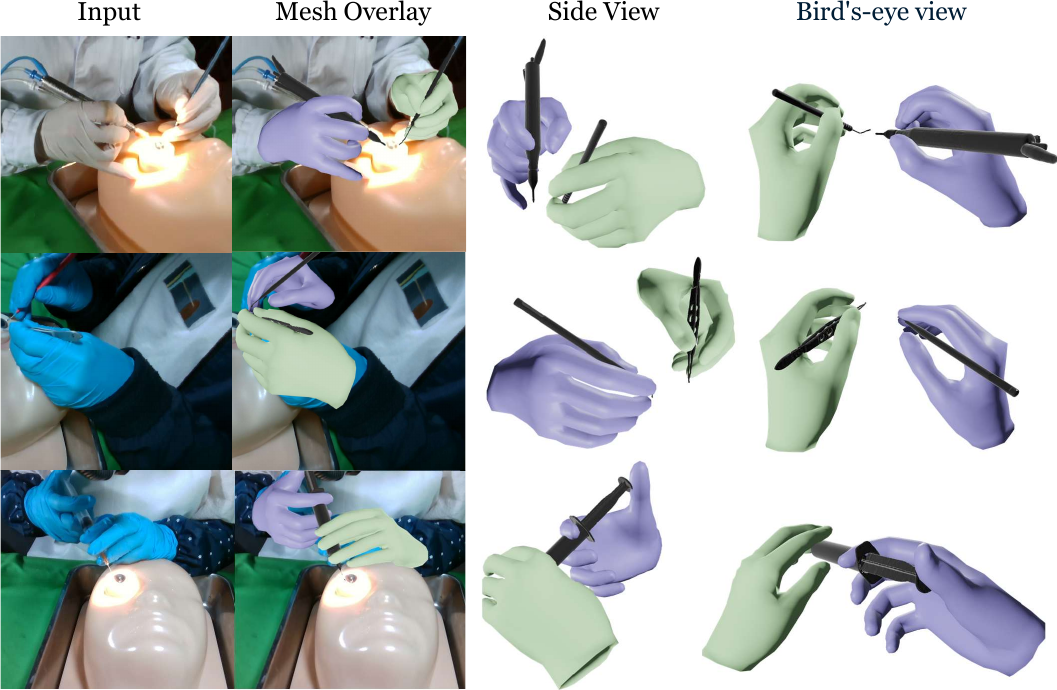}
\caption{\textbf{Qualitative results on the Hand-instrument interaction benchmark.} Each row shows a sample from the test set, with columns displaying: (1) input image, (2) mesh prediction, (3) rendered mesh from a side view, and (4) bird’s-eye view. 
}
  \label{fig:qualitative}
\end{wrapfigure}

\noindent\textbf{Results.} We evaluate the proposed OH-Net framework on the hand-instrument interaction benchmark. As shown in \cref{tab:handobj}, our method achieves state-of-the-art performance across all evaluated metrics, including MPJPE, MPVPE, MRRTE, and ADD-S. The full OH-Net benefits significantly from the joint modeling of hands and instruments, with joint training improving both interaction accuracy and articulation consistency. Notably, OH-Net is the first method capable of reconstructing two hands and two interacting instruments simultaneously, with explicit disentanglement and spatial reasoning. The RGB-D version, OH-Net-D, shows further gains across all metrics, especially in interaction-specific scores such as Penetration (Pen) and MRRTE$^{h,o}$. The MAE and ADD-S results highlight reliable articulation estimation and instrument localization. Moreover, \cref{fig:qualitative} showcases qualitative examples of predicted meshes and multi-view renderings, demonstrating our model's ability to preserve hand-instrument contact and recover detailed interactions even under occlusions. Additional results and visualizations are provided in~\ref{baselines:appendix}.


\vspace{-2mm}
\section{Discussion}
\label{sec:discussion}

\vspace{-2mm}
\noindent \textbf{Related Work.} Surgical vision research has recently evolved beyond traditional tasks~\cite{nwoye2021rendezvous, hu2024ophnet, grammatikopoulou2021cadis} toward 3D perception, including scene reconstruction~\cite{zha2023endosurf, cotsoglou2024dynamic}, depth estimation~\cite{cui2024surgical, yang2024self}, and navigation~\cite{qiu2020endoscope, yang2024slam}. Despite advances in datasets like AVOS~\cite{goodman2024analyzing} and MM-OR~\cite{ozsoy2025mm}, most systems remain limited to passive monitoring rather than real-time surgical assistance. Meanwhile, 3D hand reconstruction has progressed from general benchmarks~\cite{zimmermann2019freihand, moon2020interhand2} to synthetic medical datasets~\cite{wang2023pov, birlo2024hup}, though lacking clinical realism. Monocular approaches have evolved from single-hand models~\cite{dong2024hamba, pavlakos2024reconstructing} to two-hand systems~\cite{zhang2021interacting, lin20244dhands} and hand-object interactions~\cite{wu2024reconstructing, ye2023diffusion}, using template-based methods~\cite{garcia2018first, hampali2020honnotate} and geometric constraints~\cite{brahmbhatt2020contactpose, zhang2020perceiving}. Our work bridges the clinical gap with a targeted 3D perception framework for ophthalmic microsurgery using real RGB-D data capturing fine-grained, two-hand, multi-instrument interactions. See Tab. \ref{tab:comp_general} and~\ref{sec:related_appendix} for details.


\noindent \textbf{Conclusion.} In this paper, we introduce OphNet-3D, the first large-scale, real-world RGB-D dynamic 3D reconstruction dataset for ophthalmic microsurgery, comprising 41 sequences from 40 surgeons (7.1 M frames), annotated with 12 surgical phases, 10 instrument classes, detailed MANO hand meshes, and 6D instrument poses. We develop a multi-stage automatic annotation pipeline that integrates monocular hand-prior models, segmentation–point-cloud alignment, and biomechanics-based joint hand–instrument optimization. Leveraging this dataset, we establish two new benchmarks—bimanual hand pose estimation and hand–instrument interaction—and propose H-Net and OH-Net, which achieve state-of-the-art performance on all metrics ().

\noindent \textbf{Limitations \& Future Work.}  This study offers new insights into dynamic 3D reconstruction in ophthalmic surgery but has three main limitations: it relies on data from a single center (requiring multi-center validation for better generalizability); strong microscope illumination causes instrument-tip overexposure (which could be addressed with synchronized motion-capture RGB or infrared imaging); and it has not yet integrated the microscope view for joint reconstruction of the ocular surface, hands, and instruments (may be a key direction for future work to boost clinical relevance).

{
    \small
    \bibliographystyle{abbrv}
    \bibliography{main}
}

\newpage
\appendix

\section{Related Work}
\label{sec:related_appendix}


\noindent \textbf{Surgical 3D Perception.} 
Beyond conventional tasks in surgical assistance such as phase recognition~\cite{nwoye2021rendezvous, hu2024ophnet, hu2023nurvid, hu2024ophclip}, anatomical segmentation~\cite{borgli2020hyperkvasir, grammatikopoulou2021cadis} and instrument detection~\cite{hu2024ophclip, nwoye2021rendezvous}, recent efforts have increasingly focused on 3D perception for open surgical environments, addressing challenges in scene reconstruction~\cite{zha2023endosurf, cotsoglou2024dynamic, hayoz2024online, chen2025endoperfecthybridnerfstereovision}, depth estimation~\cite{cui2024surgical, yang2024self, tian2024endoomni, liu2019dense}, navigation~\cite{qiu2020endoscope, bartholomew2024surgical, yang2024slam, manni2024bodyslam}, and skill assessment~\cite{liu2021towards, goodman2024analyzing}. A growing body of work leverages multi-view camera arrays, RGB-D sensors, and IMUs to reconstruct dense surgical scenes with both static anatomical structures and dynamic hand-tool interactions. Goodman et al.~\cite{goodman2024analyzing} proposed AVOS, a large-scale annotated video dataset of open surgeries, and developed a multitask model to extract procedural signatures and quantify surgeon skill from real-world surgical videos. MM-OR~\cite{ozsoy2025mm} introduces a comprehensive multimodal operating room dataset featuring RGB-D, audio, speech transcripts, and robotic logs, annotated with panoptic segmentation and semantic scene graphs. While these systems demonstrate technical progress on specific tasks and in multimodal fusion capabilities, their clinical impact remains limited, with most systems designed primarily for passive action recognition or offline documentation. Few are optimized for real-time use or tailored to support specific training objectives in microsurgical procedures. In this work, we address this gap by developing a clinically-oriented 3D perception framework for ophthalmic microsurgery, integrating task-specific data, model design, and system-level considerations for real-world deployment.

\noindent \textbf{3D Hand Dataset.} 
Recent 3D hand reconstruction datasets have driven progress in hand pose, object pose, and interaction reconstruction (see Tab.~\ref{tab:comp_general}). General benchmarks such as FreiHAND~\cite{zimmermann2019freihand} and InterHand2.6M~\cite{moon2020interhand2} provide multi-view real RGB images for reliable 3D pose recovery. ContactPose~\cite{brahmbhatt2020contactpose} further leverages RGB-D to capture detailed contact patterns. Banerjee et~al. present HOT3D~\cite{banerjee2024hot3d}, a large-scale egocentric, multi-view hand--object interaction (HOI) dataset comprising $\approx$1.5\,M synchronized frames captured with head-mounted Aria and Quest~3 cameras across 19 participants manipulating 33 objects; each frame is paired with mocap-grade 3D hand and object poses. Medical datasets extend these ideas to surgery. Hein et al.~\cite{hein2021towards} released a synthetic surgical RGB set with limited views and simple actions. POV-Surgery~\cite{wang2023pov} and HUP-3D~\cite{birlo2024hup} enrich viewpoint count and interaction diversity, yet their synthetic origin still curbs realism. Ophthalmic surgery, notably cataract removal, follows fixed workflows but requires both hands to manipulate several tools in a confined field—for instance, holding an iris retractor while guiding capsulorhexis forceps. These dense, dual-hand motions strain current reconstruction methods. OphNet-3D addresses this gap with a real RGB-D clinical dataset that records fine-grained, two-hand, multi-instrument interactions during live ophthalmic operations.

\noindent \textbf{Monocular 3D Hand Mesh Reconstruction.} 
Compared with multi‑view approaches, monocular reconstruction is more practical in clinical theatres: a single, non‑contact RGB camera minimises equipment, preserves sterility, and avoids obstructing the surgeon’s workspace. Research has progressed from single‑hand models~\cite{dong2024hamba, li2024hhmr, pavlakos2024reconstructing, kim2023sampling, yu2023acr} to two‑hand systems~\cite{zhang2021interacting, yu2023acr, lin20244dhands, ren2023decoupled} and, most recently, HOI reconstruction~\cite{wu2024reconstructing, prakash20243d, huang2022hhor, ye2023diffusion}. Monocular HOI remains difficult because occlusion is severe and annotated data are scarce. Many studies therefore assume known instance‑specific templates~\cite{garcia2018first, hamer2010object, hampali2020honnotate}; with the template in place, object recovery reduces to 6D pose estimation and joint hand–object pose inference. Joint reasoning is implemented via implicit feature fusion~\cite{chen2021joint, gkioxari2018detecting, liu2021semi, shan2020understanding, tekin2019h+}, explicit geometric constraints such as contact or collision~\cite{brahmbhatt2020contactpose, cao2021reconstructing, corona2020ganhand, grady2021contactopt, zhang2020perceiving}, or physics‑based consistency~\cite{pham2017hand, tzionas2016capturing}. To lift the template assumption, newer work directly predicts object shape—either as explicit genus‑0 meshes~\cite{hasson2019learning} or via a joint hand–object implicit field~\cite{karunratanakul2020grasping}. Conditional reconstruction strategies further exploit hand joint cues to refine object geometry~\cite{ye2022s}.

\section{OphNet-3D Construction}
\subsection{Synchronized Recording Configuration}
\label{configuration}

We capture using the Intel RealSense SDK and synchronize all cameras with the official software-trigger method. Because we simultaneously acquire and store high-resolution RGB and depth streams from eight cameras, the system’s USB bandwidth and I/O performance are pushed to their limits, and sustained data throughput may exceed its capacity, causing occasional frame drops. Therefore, on the hardware side, we’ve configured a high-core-count host machine, enterprise-grade storage drives with high write speeds, and fiber-optic USB cables to ensure stable data transmission.

In high-frequency image I/O scenarios, traditional storage formats such as JPEG and PNG—despite their widespread compatibility and ease of visualization—incur compression artifacts and processing overhead that can significantly degrade overall system performance. To address these limitations and improve data throughput, we adopt a binary-format–based image encoding approach that directly serializes cv::Mat matrix data into raw binary files. Specifically, our method first writes essential matrix metadata (rows, columns, and cv::Mat type) as integers in the file header to ensure accurate reconstruction of the image’s dimensions and layout; it then appends the unaltered pixel byte stream to the file body without any compression or encoding conversion. During decoding, the file is read in the same order to restore the original cv::Mat structure exactly. In our implementation, color frames are stored as 848 × 480 three-channel matrices of type CV\_8UC3, while depth frames are 848 × 480 single-channel matrices of 16-bit unsigned integers (UINT16), with each color–depth pair spatially aligned. For acquisition, the main thread initializes eight camera-capture subthreads; upon receiving a capture command, image saving begins and the preview display is disabled to conserve resources, whereas the preview is re-enabled when capture is inactive.

\begin{figure*}[htp]
\includegraphics[width=0.95\textwidth]{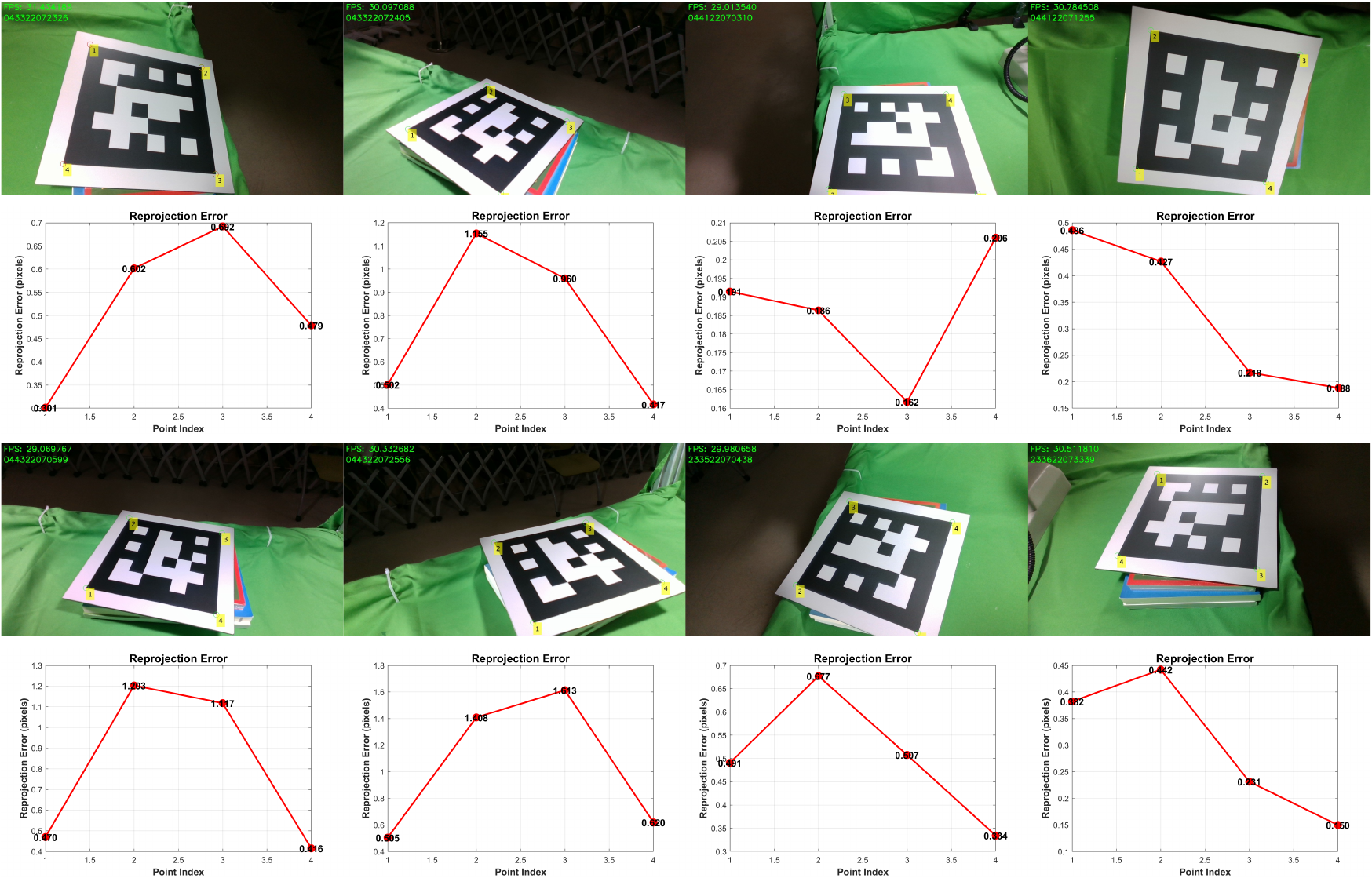}
    \centering
    \caption{Synchronized calibration of 8 cameras.}
    \label{fig:cali3}
\end{figure*}

\subsection{Synchronized Calibration}
In our multi‐camera calibration pipeline, each Intel RealSense D435’s intrinsic parameters—including focal lengths, principal point coordinates and lens distortion coefficients—are retrieved at runtime via the official SDK and assembled into the 3×3 camera matrix $K$; for extrinsic calibration, a planar ArUco marker board of precisely known marker size and layout serves as the world reference, and for each synchronized color capture a simple green‐channel mask first isolates the board region, from which the 2D corner coordinates of each detected ArUco marker are paired with their exact 3D positions on the board plane. A non‐linear Perspective‐n‐Point solver then computes the rigid‐body rotation and translation that best align the 3D marker positions to their 2D observations, yielding a 4×4 homogeneous transform from board (world) to camera coordinates. During the recording process, due to force majeure the platform was moved three times. After each move, we performed a recalibration; Fig.~\ref{fig:cali3} shows one such calibration example.


\subsection{Phase Definition and Demonstration}
\label{phase_definition}
\begin{figure*}[t!]
\includegraphics[width=1\textwidth]{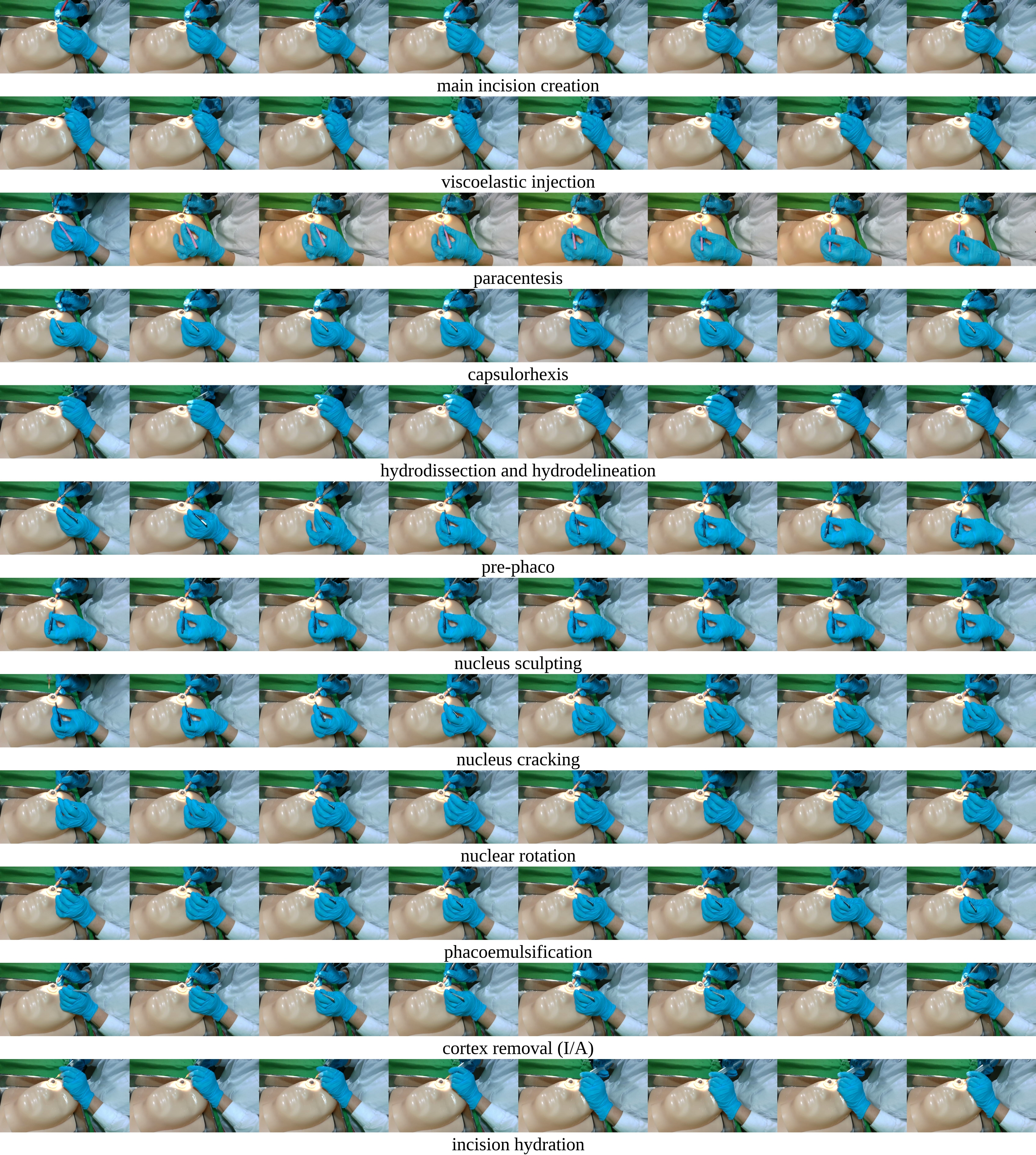}
    \centering
    \caption{12 phases from 2 different views.}
    \label{phase_pre2}
\end{figure*}

Referring to the standard cataract surgical procedure, we divided the entire surgical process into 12 phases: \emph{(1) main incision creation, (2) viscoelastic injection, (3) paracentesis, (4) capsulorhexis, (5) hydrodissection and hydrodelineation, (6) pre-phaco, (7) nucleus sculpting, (8) nucleus cracking, (9) nuclear rotation, (10) phacoemulsification, (11) cortex removal, and (12) incision hydration}. Detailed definitions of each phase and the instruments used are listed in Tab.~\ref{tab:phase_definition}. Fig.~\ref{phase_pre2} shows the 12 phases from two different camera viewpoints.

\begin{table*}[htbp]
\centering
\resizebox{0.95\textwidth}{!}{%
\begin{tabular}{c|p{3.5cm}|p{11cm}|l@{}}
\toprule
\textbf{Index} &\textbf{Phase Label} & \textbf{Phase Definition} & \textbf{Instruments Used} \\ \midrule
1 & main incision creation & Use keratome to create a precise entry point through the cornea or  limbus to access the anterior chamber of the eye. & toothed forceps, keratome blade \\  \midrule

2 &viscoelastic injection & Injection of ophthalmic viscoelastic devices (OVDs) into the anterior chamber using a viscoelastic syringe to maintain chamber depth, protect intraocular tissues, and facilitate subsequent surgical steps. & viscoelastic syringe \\  \midrule

3 &paracentesis & Creation of a small, self-sealing side-port incision at the limbus using a 15° stab blade to allow access for second instruments such as the chopper into the anterior chamber. &   15° stab blade, toothed forceps\\  \midrule

4 &capsulorhexis & Creating a continuous curvilinear opening in the anterior lens capsule using a capsulorhexis forceps to allow safe access to the lens nucleus for removal. & iris repositor, capsulorhexis forceps \\  \midrule

5 &hydrodissection and hydrodelineation & Inject balanced salt solution (BSS) using a 10ml syringe to separate the lens nucleus from the cortex and capsule, allowing easier rotation and removal. & 10 mL syringe \\  \midrule

6 &pre-phaco & Use a phacoemulsfication handpiece gently remove loose cortical or epinuclear material from the anterior lens surface before nucleus sculpting begins. & nucleus chopper, phacoemulsification handpiece \\  \midrule

7 &nucleus sculpting & Using a phacoemulsification handpiece to carve grooves into the lens nucleus to facilitate nucleus division and removal. & nucleus chopper, phacoemulsification handpiece \\  \midrule

8 &nucleus cracking & Using the phacoemulsification handpiece in combination with a nucleus chopper to mechanically split the grooved nucleus into smaller fragments for easier phacoemulsification and removal. & nucleus chopper, phacoemulsification handpiece \\  \midrule

9 &nuclear rotation & Using a nucleus chopper to gently rotate the lens nucleus within the capsular bag, ensuring optimal positioning for continued phacoemulsification. & nucleus chopper, phacoemulsification handpiece \\  \midrule

10 &phacoemulsification & Involves using a phacoemulsification handpiece with ultrasonic tip to break up and emulsify the nucleus fragments, while simultaneously aspirating the lens material and maintaining anterior chamber stability. & nucleus chopper, phacoemulsification handpiece \\  \midrule

11 &cortex removal (I/A) & Use an irrigation-aspiration handpiece to gently remove the residual cortical material from the capsular  bag following phacoemulsification of the lens nucleus. & iris repositor, irrigation-aspiration handpiece \\  \midrule

12 &incision hydration & Using a balanced salt solution (BSS) through a 10mL syringe to swell the corneal stroma at the incision (both main incision and paracentesis) edges, sealing the surgical incisions at the end of cataract surgery. &  10 mL syringe \\  \bottomrule

\end{tabular}%
}
\caption{Definition of each phase and the instruments used in each phase}
\label{tab:phase_definition}
\end{table*}

\subsection{Phase Localization Annotation}
\label{phase_localization}
We performed phase–boundary annotation on each video sequence as follows. First, because the microscope and hand‐view recordings were acquired on separate devices and thus lack intrinsic temporal synchronization, we aligned them manually. Immediately before each trial, a rigid printed marker bearing the legend “Start Recording” was displayed concurrently across all camera views; the frame in which this marker first appeared in each view served as the synchronization point. Next, an experienced ophthalmologist reviewed the temporally aligned microscope and hand‐view videos to delineate the onset and offset of each surgical phase. To improve label purity, segments corrupted by visual noise—such as instrument exchanges during which the hand left the camera field of view—were excluded. A second ophthalmologist then independently verified and revised all boundary annotations. Finally, to ensure precise demarcation of phase transitions, we uniformly contracted each annotated interval by removing one second from both its start and end.

\subsection{Instrument Demonstration}
\label{instrument}
\begin{figure*}[htp]
\includegraphics[width=0.85\textwidth]{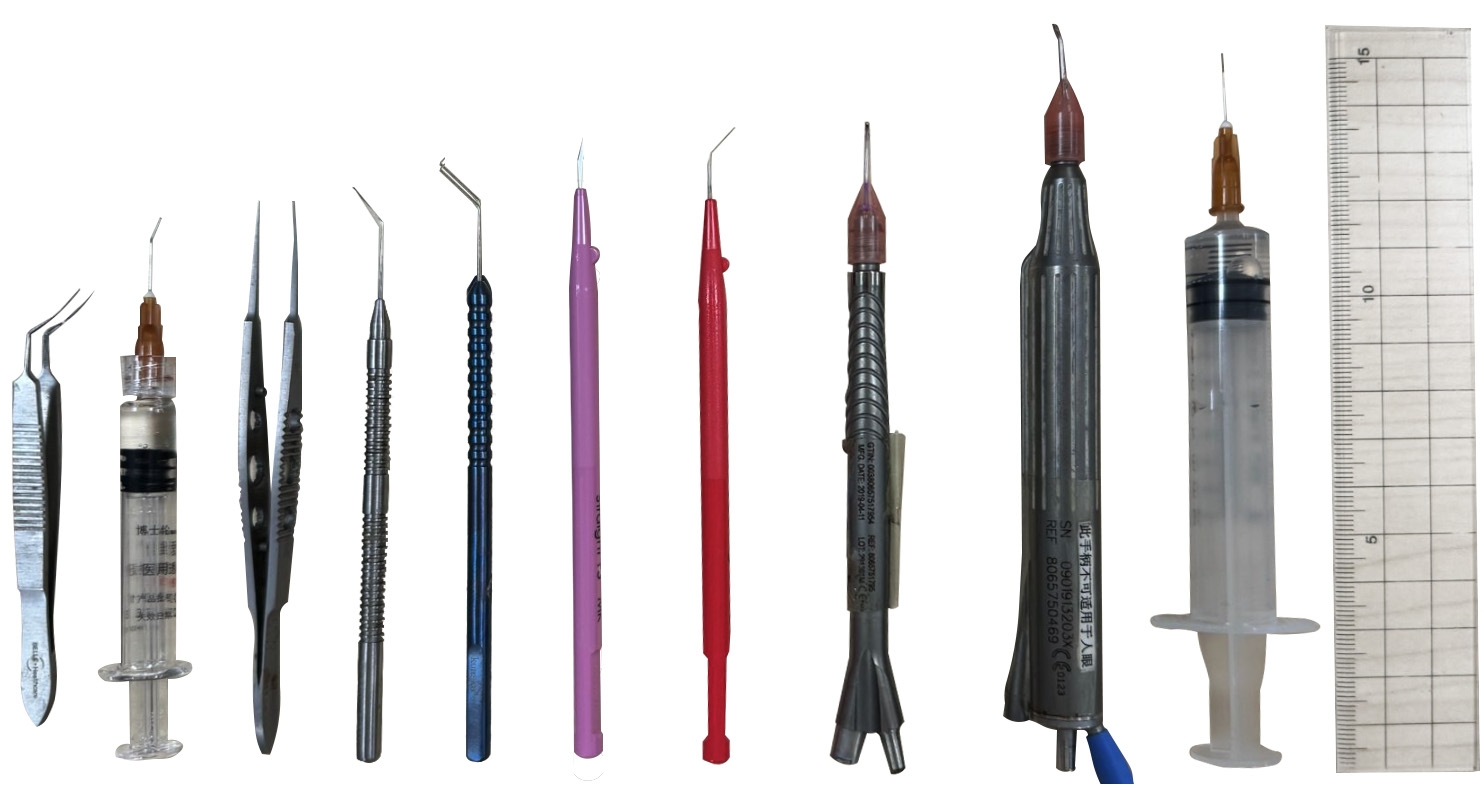}
    \centering
    \caption{10 different instruments. From left to right they are: capsulorhexis forceps, viscoelastic syringe, toothed forceps, iris repositor, nucleus chopper,  15° stab blade, keratome blade, irrigation-aspiration handpiece, phacoemulsification handpiece, and 10 mL syringe.}
    \label{fig:instrument}
\end{figure*}

\begin{figure*}[t!]
    \begin{subfigure}[t]{0.49\textwidth}
        \includegraphics[width=\textwidth]{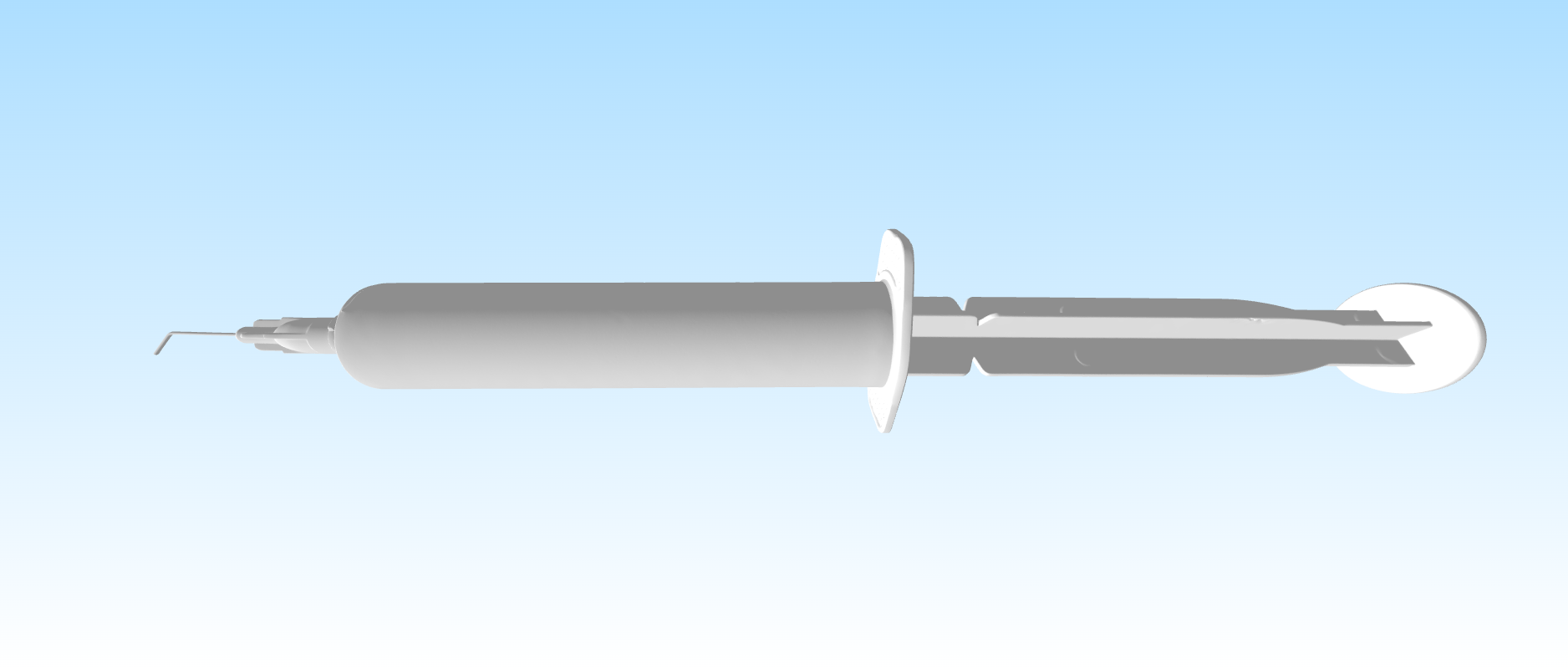}
        \vspace{-3pt}
        \includegraphics[width=\textwidth]{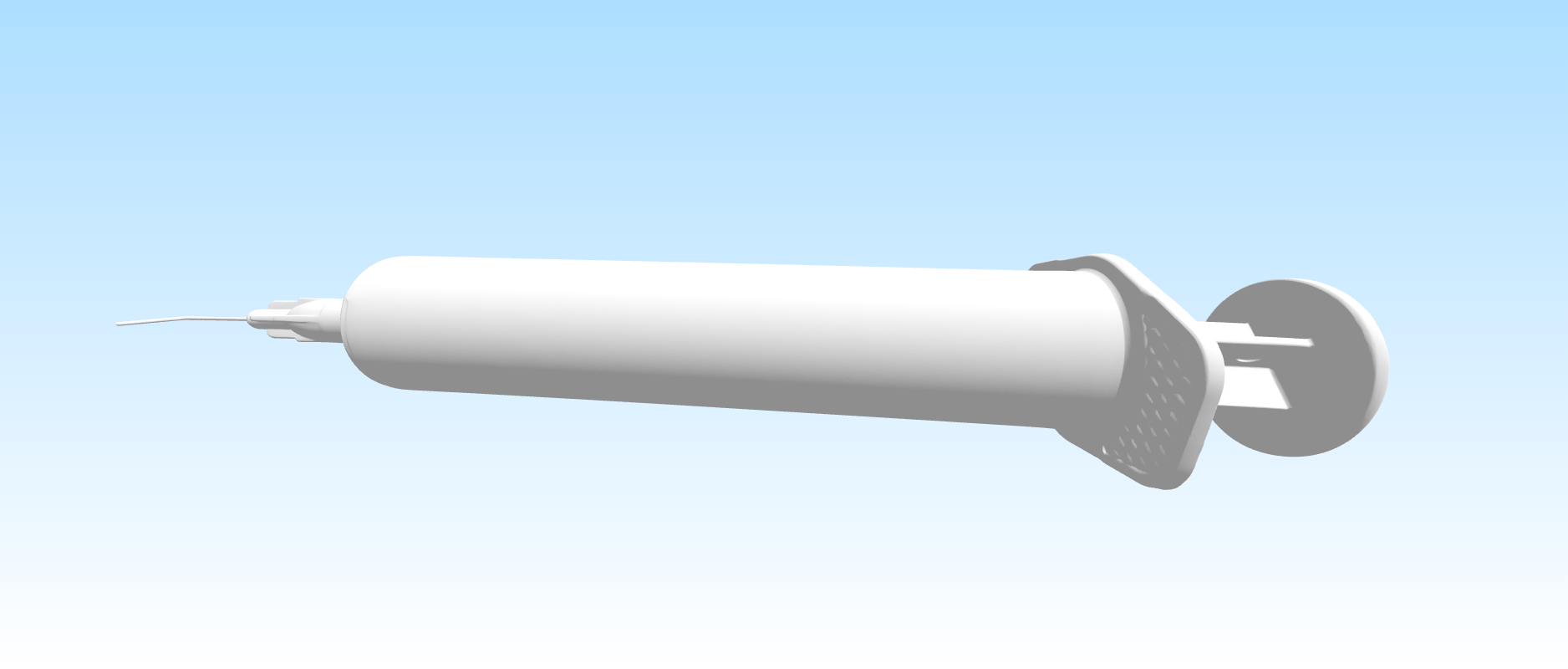}
        \vspace{-3pt}
        \includegraphics[width=\textwidth]{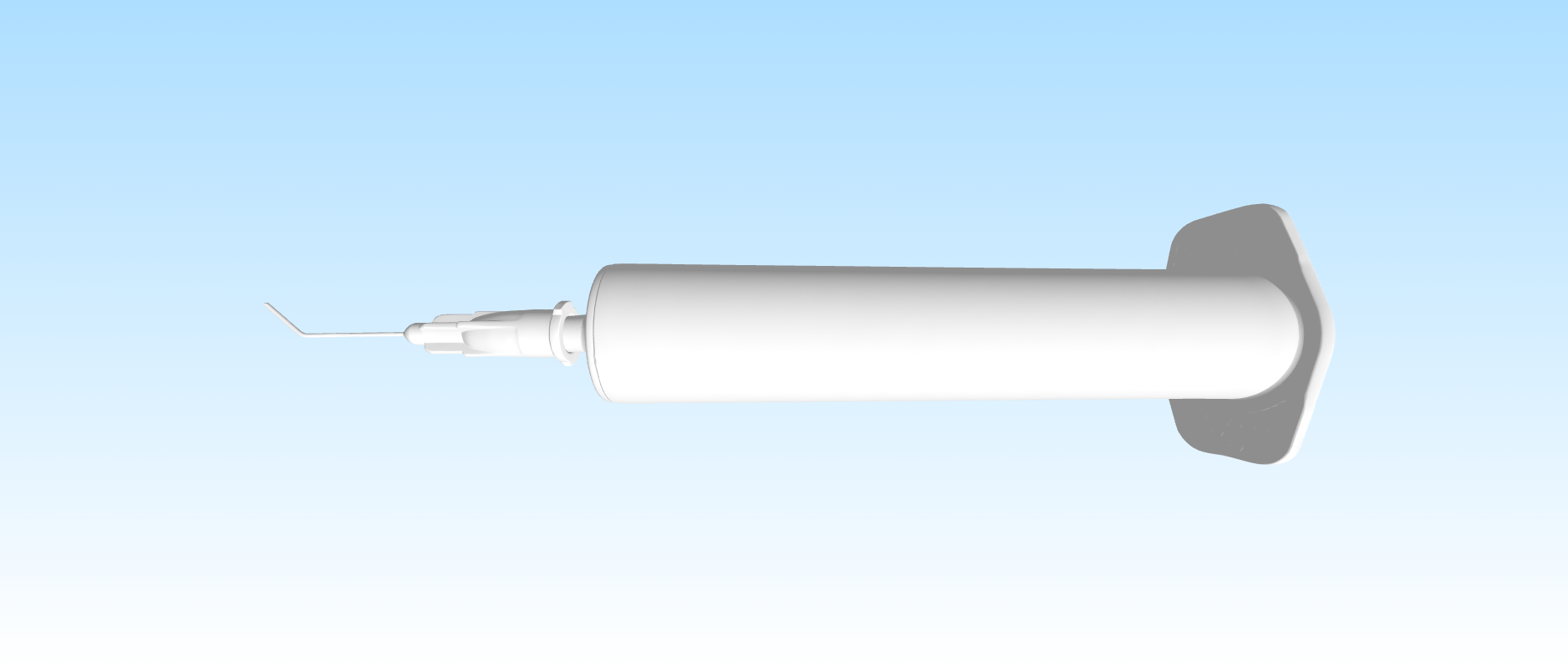}
        \vspace{-3pt}
        \includegraphics[width=\textwidth]{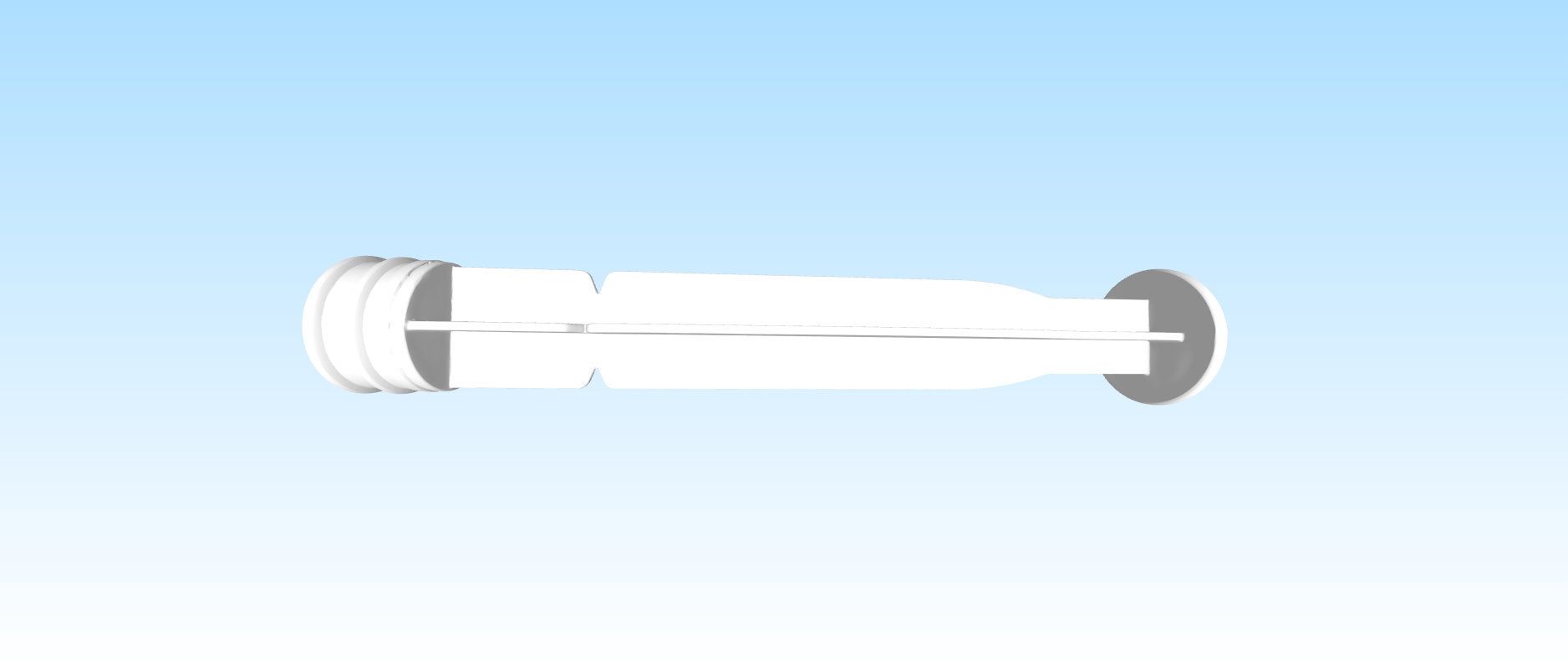}
        \caption{10 mL syringe}
        \label{paltform_I}
    \end{subfigure}
    \hfill
    \begin{subfigure}[t]{0.49\textwidth}
        \includegraphics[width=\textwidth]{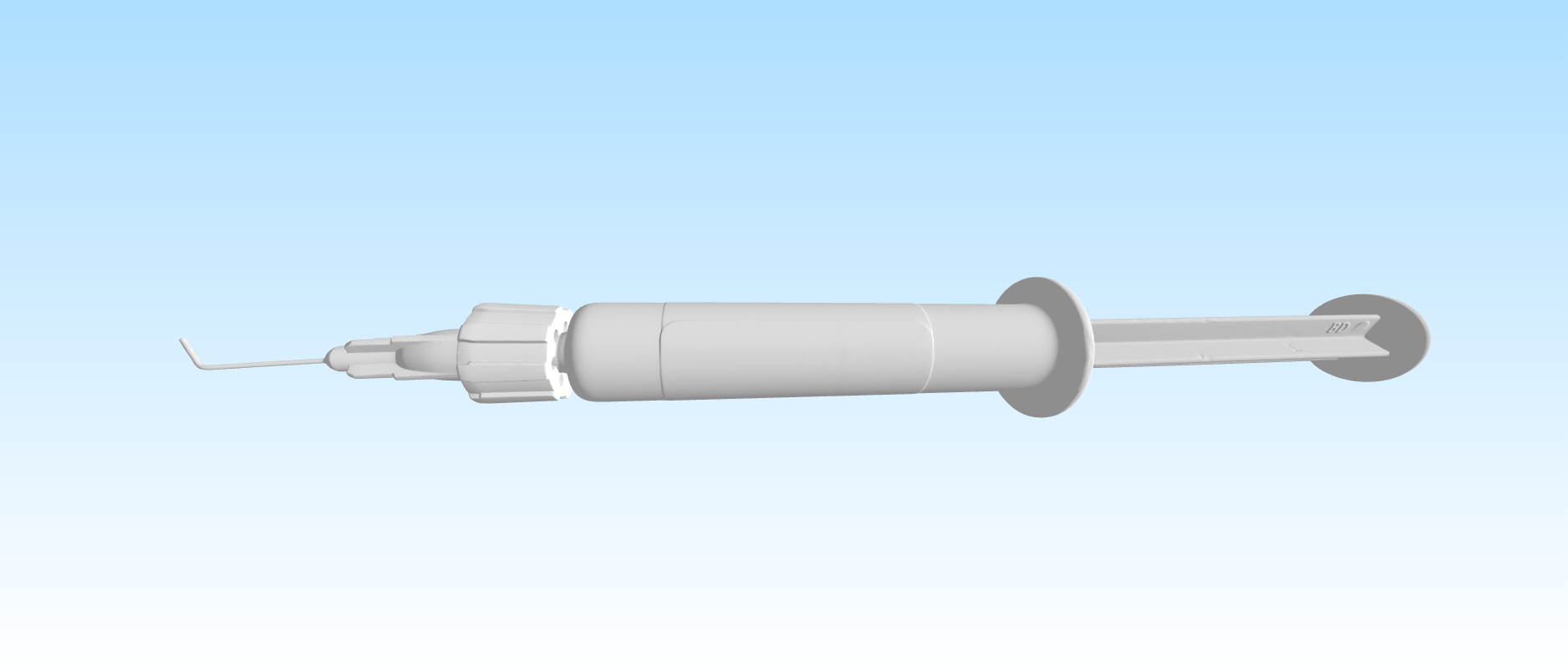}
        \vspace{-3pt}
        \includegraphics[width=\textwidth]{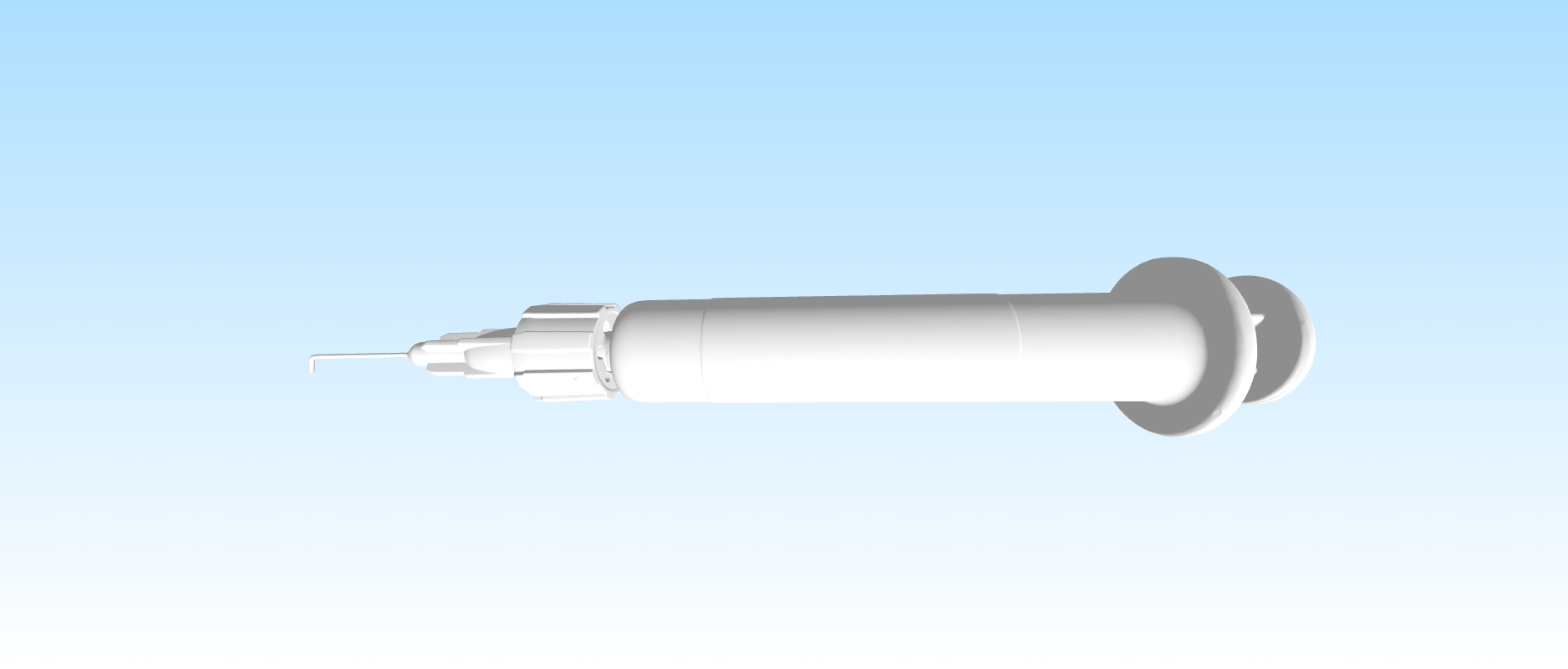}
        \vspace{-3pt}
        \includegraphics[width=\textwidth]{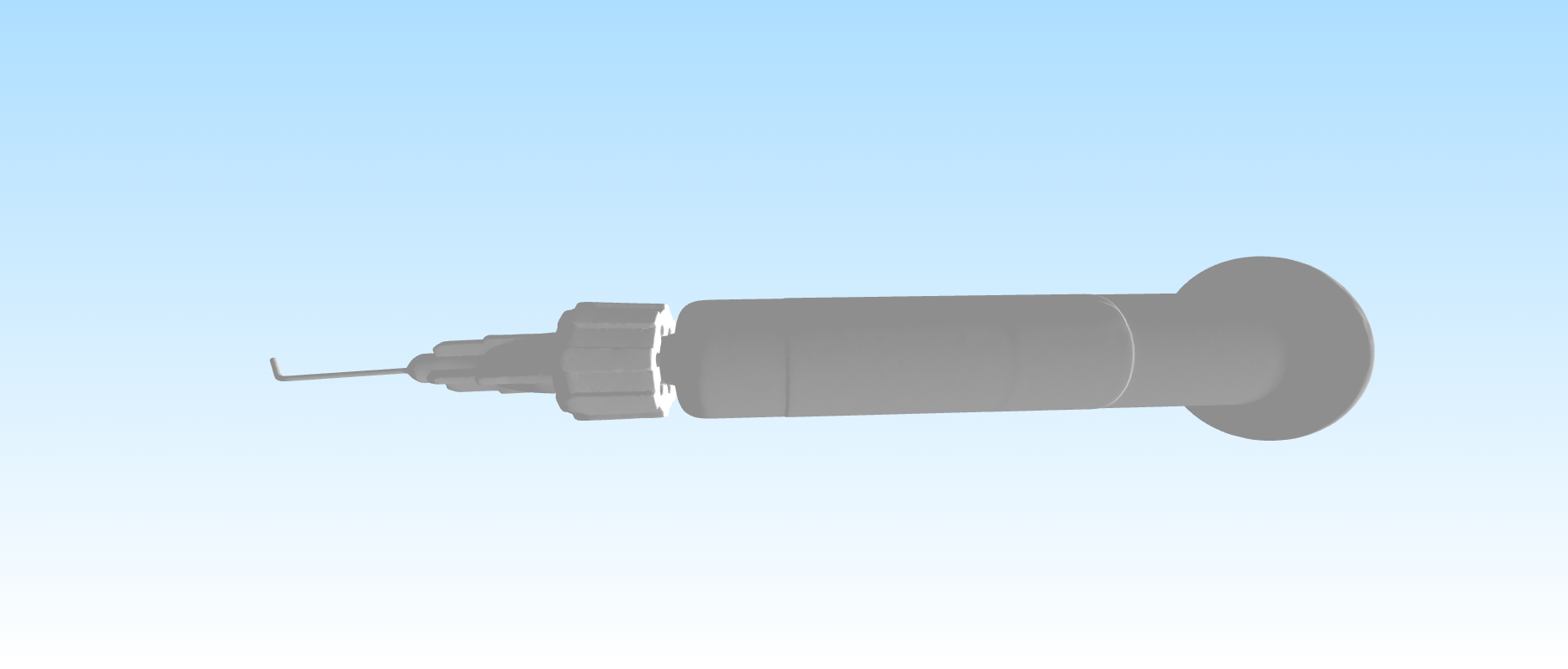}
        \vspace{-3pt}
        \includegraphics[width=\textwidth]{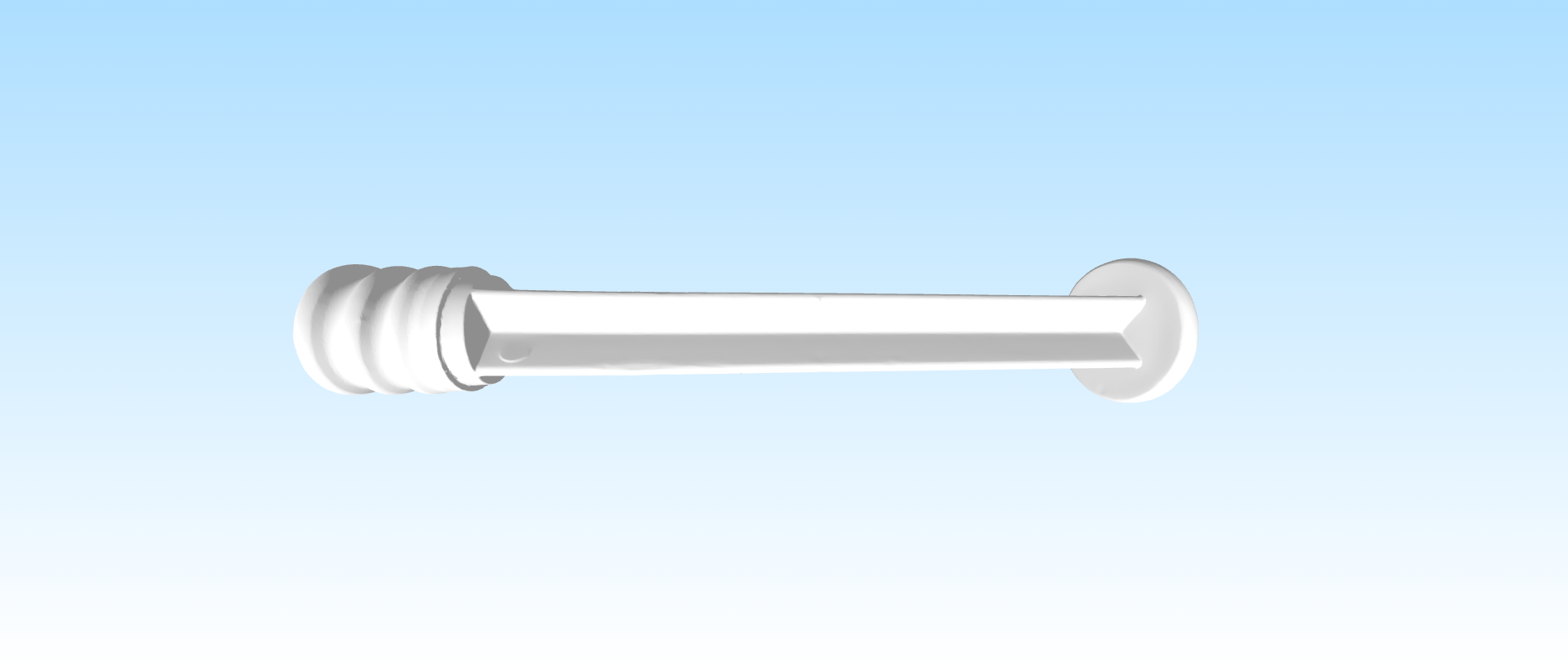}
        \caption{viscoelastic syringe}
        \label{paltform_J}
    \end{subfigure}
    \caption{Scanned model files for 10 types of instruments.}
    \label{fig:instrument_3d}
\end{figure*}

During the procedures, 10 different surgical instruments were employed: \emph{(1) 15° stab blade, (2) keratome blade, (3) iris repositor, (4) nucleus chopper, (5) toothed forceps, (6) capsulorhexis forceps, (7) phacoemulsification handpiece, (8) irrigation-aspiration handpiece, (9) 10 mL syringe, and (10) viscoelastic syringe}. Fig.~\ref{fig:instrument} shows photographs of the ten instruments used during surgery, and Fig.~\ref{fig:instrument_3d} presents the 3D model scan files. For the forceps, we scanned both the open and closed configurations. For the syringe, we scanned two states—plunger rod at its maximum and minimum extension—and additionally scanned the plunger rod and the syringe body separately as individual components.



\section{OphNet-3D Statistics}
Tab.~\ref{tab:data_split_supp} reports, for each split of OphNet-3D, the number of video clips and the distribution of frame counts per surgical phase. Since some phase operations may involve a third hand entering the scene, we remove those segments during annotation; this can truncate a complete phase into multiple shorter clips. Such clips are counted separately in our statistics but are linked via sequential index identifiers.

\begin{table*}[t!]
\centering
\tiny
\resizebox{0.95\textwidth}{!}{%
\begin{tabular}{l|cccc|cccc}
\toprule
\multirow{2}{*}{\textbf{Phase Label}} & \multicolumn{4}{c|}{\textbf{No. of Clips}} & \multicolumn{4}{c}{\textbf{No. of Frames}} \\
\cmidrule(lr){2-5} \cmidrule(lr){6-9}
 & \textbf{train} & \textbf{val} & \textbf{test} & \textbf{all} & \textbf{train} & \textbf{val} & \textbf{test} & \textbf{all} \\ \midrule
main incision creation & 32 & 3 & 8 & 43 & 312,224 & 21,600 & 48,960 & 382,784\\ \midrule
viscoelastic injection & 59 & 8 & 20 & 87 & 519,120 & 59,040 & 132,960 & 711,120\\ \midrule
paracentesis & 35 &  4& 13 & 52 & 196,080 & 13,440 & 71,520 & 281,040 \\ \midrule
capsulorhexis & 49 & 4 & 15 & 68 & 1,116,960 & 105,360 & 292,080 & 1,514,400\\ \midrule
hydrodissection and hydrodelineation & 30 & 4 & 8 & 42 & 321,864 & 21,120 & 79,920 & 422,904\\ \midrule
pre-phaco & 31 & 4 & 8 & 43 & 435,600 & 48,960 & 99,600 & 584,160 \\ \midrule
nucleus sculpting & 30 & 3 &  8& 41 & 367,920 & 20,400 & 106,080 & 494,400 \\ \midrule
nucleus cracking & 29 & 3 & 8 & 40 & 127,200 & 8,160 & 39,840 & 175,200 \\ \midrule
nuclear rotation & 12 & 1 & 5 &  18 & 51,120 & 2,640 & 31,440 & 85,200\\ \midrule
phacoemulsification & 29 & 4 & 8 & 41 & 556,080 & 46,320 & 237,600 & 840,000 \\ \midrule
cortex removal (I/A) & 33 & 3 & 10 & 46 & 751,832 & 98,640 & 290,640 & 1,141,112\\ \midrule
incision hydration & 32 & 4 & 8 & 44 & 406,168 & 22,560 & 80,672 & 509,400\\ \midrule
all & 401 & 45 & 119 & 565 & 4,955,272 & 468,240 & 1,511,312 & 7,141,720 \\ \bottomrule
\end{tabular}%
}
\caption{Phase distribution across splits for clips and frames.}
\label{tab:data_split_supp}
\end{table*}



\section{More Implementation Details and Results}\label{experiment:appendix}

\subsection{Data processing}\label{processing:appendix}
\paragraph{Hand Mesh Initialization.}
To initialize the per-frame hand motion for each view, we adopt a three-stage process inspired by DynHaMR~\cite{yu2024dyn}, incorporating 2D detection fusion, per-view tracking, and global fusion.

We first extract 2D hand keypoints by applying ViTPose~\cite{xu2022vitpose} to each RGB frame. Keypoints below a confidence threshold of $\epsilon_j = 0.5$ are discarded. Cropped patches around detected hand regions are then reprocessed through ViTPose for local refinement. To address unreliable or missing joints, we additionally apply MediaPipe~\cite{lugaresi2019mediapipe} and fuse the results: for each joint, we retain the ViTPose prediction if above threshold, and replace it with MediaPipe’s output otherwise. If entire hands are undetected in some views, we infill the missing detections by copying hand motion from nearby frames with high visibility and smoothing their trajectory with a temporal window. This approach ensures a full joint set $\hat{\mathbf{J}}^h_t$ per frame for each view. To reduce noise from hallucinated detections or wong handedness, we adopt a filtering strategy. For each frame, we keep only the bounding box with the highest IoU ($>0.9$) among overlapping detections and discard those that appear in fewer than 10 frames across the sequence. Additionally, we track bounding box continuity to detect erroneous handedness flips or duplicated hands — if a bounding box IoU with the previous frame drops below 0.1, we mark the frame as invalid and exclude it from subsequent fitting. These invalid frames are later recovered via generative infilling.

Next, we estimate the 3D hand pose per view using the coarse-to-fine regression pipeline of~\cite{pavlakos2024reconstructing}, which returns MANO parameters \(\{\pose^h_t, \shape^h_t, \orient^h_t, \trans^h_t\}\). The 3D wrist translation $\trans^h_t$ is obtained via depth sampling and back-projection:
\[
x = \frac{z(u - c_x)}{f_x}, \quad y = \frac{z(v - c_y)}{f_y},
\]
where $(u,v)$ are 2D keypoints, and $(f_x, f_y)$ are focal lengths. We choose the optimal $z$ minimizing reprojection error to initialize depth.

Finally, per-view MANO parameters are transformed to world coordinates using known camera extrinsics \(\{\mathbf{R}_i, \mathbf{t}_i\}\), and merged across views using a weighted average. View weights are derived from the per-frame visibility scores computed from 2D keypoint confidence. This results in a globally consistent, temporally smooth initialization of hand pose across all frames and views. The resulting motion serves as the input to our multi-stage RGB-D optimization pipeline \cref{sec:instrument_annotation}, where temporal, geometric, and interaction constraints are jointly optimized.

\subsection{Annotation pipeline}\label{pipeline:appendix}

\paragraph{Implementation details.} We implement the annotation pipelien with PyTorch \cite{paszke2019pytorch}. During the optimization of stage II and stage III (\cref{sec:instrument_annotation}), we use L-BFGS algorithm with $lr=1$ and optimizing the loss functions using below weights:
\begin{itemize}
    \item For stage II, we have: $\lambda_{2d}=0.001, \lambda_{\mathrm{smooth}}=10, \lambda_{\pose}=0.04, \lambda_{\shape}=0.05$.
    \item For stage III, we have: $\lambda_{z} = 200,\lambda_{\phi} = 2, \lambda_{\gamma} = 10, \lambda_{pen}=10, \lambda_{\beta}=0.05, \lambda_{\mathrm{ja}}=1, \lambda_{\mathrm{palm}}=1, \lambda_{\mathrm{bl}}=1$.
\end{itemize}
To better model the hand plausibility, we propose to leverage a biomechanical constraints and an angle limitation constraint to our objective function:
\begin{align}
    \Lja = &\sum_{j}{d_{\alpha,H}}(\bm{\alpha}_{t}^{j}, \Hb^{j}), \quad \Lbl = \sum_{j}\mathcal{I}(\|\bones^{j}_{t}\|_{2}; b_{\mathrm{min}}^{j}, b_{\mathrm{max}}^{j}),\\
    \Lpalm = &\sum_{j} \mathcal{I}(\|\curvs^{j}_{t}\|_{2}; c_{\mathrm{min}}^{j}, c_{\mathrm{max}}^{j}) + \sum_{j} \mathcal{I}(\|\dang^{j}_{t}\|_{2}; d_{\mathrm{min}}^{j}, d_{\mathrm{max}}^{j}), \\
    \Langle  =& \|\hat{\theta}_{t}^{h}\|_{2} + \mathcal{I}(\|\hat{\theta}^{h}_{t}\|_{2}; \theta_{\mathrm{min}}^{h}, \theta_{\mathrm{max}}^{h}) + \mathcal{I}(\| \hat{\theta}^{b}_{t}\|_{2}; \theta_{\mathrm{min}}^{b}, \theta_{\mathrm{max}}^{b}),
\end{align}
where $j$ is the index of the hand joint. $\Lja$ is for joint angle priors that constrains the joint angle sequence $\bm{\alpha}^{j}_{t} = (\bm{\alpha}^{f}_{t}, \bm{\alpha}^{a}_{t})$ by approximating the convex hull on $(\bm{\alpha}^{f}_{t},\bm{\alpha}^{a}_{t})$ plane with the point set $\Hb^j$. $\mathcal{I(\cdot)}$ is the interval loss that penalizes outliers. $\Lbl$ represents the loss term for bone length penalizing the finger bone length $b_{j}$ that lie outside valid bone length range $[b^{j}_\mathrm{min}, b^{j}_\mathrm{max}]$. Similarly, we further constrain the curvature $\|\curvs^{j}_{t}\|_{2}$ and angular distance $\|\dang^{j}_{t}\|_{2}$ for the palm root bones, $\Lpalm$ by penalizing the outliers if the ranges $[c_{\mathrm{min}}^{j}, c_{\mathrm{max}}^{j}]$ and $[d_{\mathrm{min}}^{j}, d_{\mathrm{max}}^{j}]$. Moreover, we constrain a specific subset of hand poses $\hat{\theta}_{t}^{h}$ (\eg twist rotation of Distal Interphalangeal (DIP) joints) and penalize the outliers of the pre-defined range $[\theta_{\mathrm{min}}^{b}, \theta_{\mathrm{max}}^{b}]$.

\paragraph{MANO Regularization.}
We regularize the predicted MANO parameters during optimization using a prior on both pose and shape:
\begin{equation}
\mathcal{L}_{\text{mano}} = \lambda_{\theta} \mathcal{L}_{\pose} + \lambda_{\beta} \mathcal{L}_{\shape}.
\end{equation}
The pose regularization term $\mathcal{L}_{\pose}$ penalizes deviations from a rest pose (assumed to be all-zero) using an $\ell_2$ norm over the pose parameters:
\begin{equation}
\mathcal{L}_{\pose} = \sum_{h \in \{l, r\}} \sum_{t=0}^{T} \| \pose^h_t \|^2_2.
\end{equation}
The shape prior $\mathcal{L}_{\shape}$ similarly penalizes the shape coefficients $\shape^h_t$, encouraging plausible hand geometry:
\begin{equation}
\mathcal{L}_{\shape} = \sum_{h \in \{l, r\}} \| \shape^h \|^2_2.
\end{equation}
These terms serve as soft constraints that prevent drift during optimization and help enforce physical realism.

\paragraph{Interaction Loss.}
To model physical plausibility and guide the relative spatial arrangement of the hand and tool, we incorporate an interaction loss $\mathcal{L}_{\mathrm{inter}}$ comprising an attraction loss $\mathcal{L}_\mathrm{A}$ and a repulsion loss $\mathcal{L}_\mathrm{R}$, following prior work~\cite{hasson2019learning}. Specifically, the attraction term encourages contact between hand and object surfaces when interaction is expected, while the repulsion term penalizes interpenetration. We define the set of hand contact vertices $\mathcal{C}^{h}_{\mathrm{ext}}$ by computing the proximity of each MANO vertex on the hand to the object mesh. For each frame, we mark as contact those hand vertices within $5\,\mathrm{mm}$ of the object surface. From this set, we extract six regions of contact based on anatomical structure: five fingertips and the palm base, following~\cite{hasson2019learning}. These six anchor regions provide soft guidance for maintaining realistic contact.

The attraction loss is computed from the set of anchor points on the hand and their closest points on the object mesh $V_\mathrm{obj}$:
\begin{equation}
\mathcal{L}_\mathrm{A}(V_\mathrm{obj}, V_\mathrm{hand}) = \sum_{i=1}^{6} \mathcal{L}_{\mathrm{dist}}(d(C^{h}_i(\mathrm{Ext}(\mathrm{Obj})), V_\mathrm{obj})),
\end{equation}
where $C^{h}_i(\mathrm{Ext}(\mathrm{Obj}))$ are the hand anchor vertices corresponding to region $i$, and $\mathcal{L}_{\mathrm{dist}}$ is a distance loss (e.g., L1) between those vertices and the nearest points on the object. We define the distance loss $\mathcal{L}_{\mathrm{dist}}$ as the average Euclidean distance between a set of hand contact vertices and their nearest neighbors on the object mesh:
\begin{equation}
\mathcal{L}_{\mathrm{dist}}(C, V_\mathrm{obj}) = \frac{1}{|C|} \sum_{\mathbf{v} \in C} \min_{\mathbf{u} \in V_\mathrm{obj}} \|\mathbf{v} - \mathbf{u}\|_2,
\end{equation}
where $C$ is the set of contact vertices on the hand and $V_\mathrm{obj}$ is the object mesh. To discourage unnatural interpenetration, we define a repulsion loss $\mathcal{L}_\mathrm{R}$ between the hand vertices and the inside of the object:
\begin{equation}
\mathcal{L}_\mathrm{R}(V_\mathrm{obj}, V_\mathrm{hand}) = \sum_{\mathbf{v}_i \in V_\mathrm{hand}} \mathbb{1}_{\mathrm{in}}(\mathbf{v}_i) \cdot d(\mathbf{v}_i, V_\mathrm{obj}),
\end{equation}
where $\mathbb{1}_{\mathrm{in}}(\cdot)$ is an indicator function marking vertices that lie inside the object, and $d(\cdot, V_\mathrm{obj})$ is the shortest distance to the object surface. The final interaction loss is a weighted combination:
\begin{equation}
\Linter = \lambda_\mathrm{R} \mathcal{L}_\mathrm{R} + (1 - \lambda_\mathrm{R}) \mathcal{L}_\mathrm{A},
\end{equation}
where $\lambda_\mathrm{R}=0.5$ balances repulsion and attraction. Following~\cite{hasson2019learning}, we empirically set $\lambda_\mathrm{R} = 1.0$ during early training to resolve interpenetration first, then reduce it to allow attraction.

This loss encourages anatomically plausible contact while suppressing mesh collisions, improving the realism of hand-tool interaction.

\paragraph{Signed Distance Field Loss $\Lsdf$.}
To penalize interpenetration between the hand mesh and the tool surface, we adopt a signed distance field loss that queries the SDF defined over the tool volume. For each time step, we precompute a voxelized signed distance field $\phi^{o}_{t}(\cdot)$ around the tool mesh $O^{h}_{t}$. Then, the SDF loss is defined as:
\begin{equation}
\Lsdf = \sum_{\mathbf{v} \in V^h_t} \max(0, -\phi^{o}_{t}(\mathbf{v}))^2,
\end{equation}
where $V^h_t$ is the hand mesh and $\phi^{o}_{t}(\mathbf{v})$ returns the signed distance of a hand vertex $\mathbf{v}$ to the tool surface — negative values indicate penetration. The $\max(0, \cdot)^2$ term ensures that only intrusions (i.e., where $\phi < 0$) are penalized, encouraging the hand to remain outside the tool surface.

\paragraph{Runtime} Our network is agnostic to the initialization method and is not restricted to using \cite{pavlakos2024reconstructing}, which affects the processing time. Therefore we conduct runtime experiment excluding the stage I (hand and instance mask initialization time) and Pyrender offscreen rendering, which are not included in the optimization pipeline. On an NVIDIA A100 GPU the optimization pipeline can take 15 minutes for a video with 1000 frames, The stage II (the initialization of instrument 6D pose and hand pose optimization) takes around 10.9 minutes to process due to the bottle neck in ICP processing and registration. Finally, the last stage of joint optimization only takes around 4.1 minutes.

\subsection{Baselines and Experiments}\label{baselines:appendix}

\paragraph{Implementation details.} We implement our baseline methods based on PyTorch \cite{paszke2019pytorch}. We use ResNet-50 \cite{he2016deep} as the backbone network. All the input image and segmentation maps are resized to 512$\times$512 while keeping the same aspect ratio with 0 paddings, which are then used to extract the feature maps $f \in \mathcal{R}^{(D+2) \times \mathrm{H} \times \mathrm{W}}$ with CoordConv \cite{}.  We train our network using 1 A100 GPU with batchsize of 64. The size of our backbone feature is $128 \times 128$ and the size of our 4 pixel-aligned output maps is $64 \times 64$. We applied random scale, rotation, flip, and colour jitter augmentation during training.

\paragraph{Loss functions.}
We supervise our baseline models using a weighted sum of losses that account for 2D keypoint projection, 3D reconstruction accuracy, silhouette alignment, segmentation, and parameter regression. The total loss is formulated as:
\begin{equation}
    \mathcal{L} = \lambda_\mathrm{focal} \mathcal{L}_{\mathrm{focal}} + \lambda_\mathrm{pj2d} \mathcal{L}_{\mathrm{pj2D}} + \lambda_\mathrm{3d} \LThreeD + \lambda_\mathrm{sil} \mathcal{L}_{\mathrm{sil}} + \mathcal{L}_{\mathrm{MANO}} + \lambda_\mathrm{seg} \mathcal{L}_{\mathrm{seg}} + \lambda_\mathrm{obj} \mathcal{L}_{\mathrm{tool}}.
\end{equation}

$\mathcal{L}_{\mathrm{focal}}$ is focal loss \cite{lin2017focal} used to supervise the predicted hand and object center heatmaps. The projection loss $\mathcal{L}_{\mathrm{pj2D}}$ penalizes the re-projection error between the predicted 3D keypoints (via MANO) and the 2D annotations using a robust Geman-McClure function. The 3D loss $\LThreeD$ measures the vertex-to-surface distance between the predicted mesh and the observed point cloud from multi-view fusion. The $\mathcal{L}_{\mathrm{pj2D}}$ and $\LThreeD$ are also computed for the instrument vertex and pre-defined 3D bounding box around it. The silhouette loss $\mathcal{L}_{\mathrm{sil}}$ compares the predicted hand and tool silhouette masks (from a differentiable renderer) against the ground-truth masks to enforce pixel-wise consistency. The MANO loss $\Lmano$ is composed of L2 losses over the predicted hand pose and shape parameters:
\begin{equation}
    \Lmano = \lambda_{\theta} \| \pose - \pose^* \|_2^2 + \lambda_{\beta} \| \shape - \shape^* \|_2^2,
\end{equation}
where $\pose^*$ and $\shape^*$ denote pseudo ground-truth values from the annotation pipeline. The segmentation loss $\mathcal{L}_{\mathrm{seg}}$ is a pixel-wise cross-entropy loss over the hand and tool instance masks. The tool loss $\mathcal{L}_{\mathrm{tool}}$ supervises both the 6D pose and the 1D articulation parameters via parameter map regression and point cloud alignment. We use the following weights in all experiments: $\lambda_\mathrm{focal}=80$, $\lambda_\mathrm{pj2d}=400$, $\lambda_\mathrm{3d}=300$, $\lambda_\mathrm{sil}=50$, $\lambda_{\theta}=80$, $\lambda_{\beta}=10$, $\lambda_\mathrm{seg}=160$.

\begin{figure}[th]
       \centering
       \includegraphics[width=\linewidth]{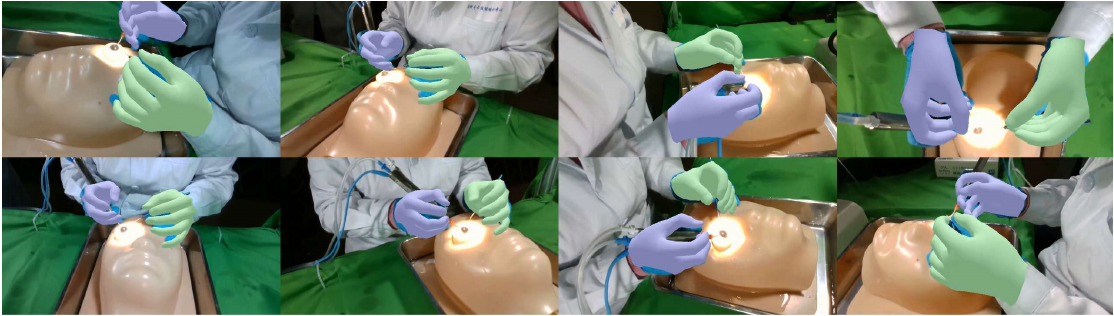}
       \caption{\textbf{Qualitative results on the hand pose estimation benchmark.} Each image is an overlay from each camera view.\vspace{-4mm}}
       \label{fig:hand:appendix} 
\end{figure}

\subsection{More Qualitative Results}
\begin{figure}[th]
       \centering
       \includegraphics[width=\linewidth]{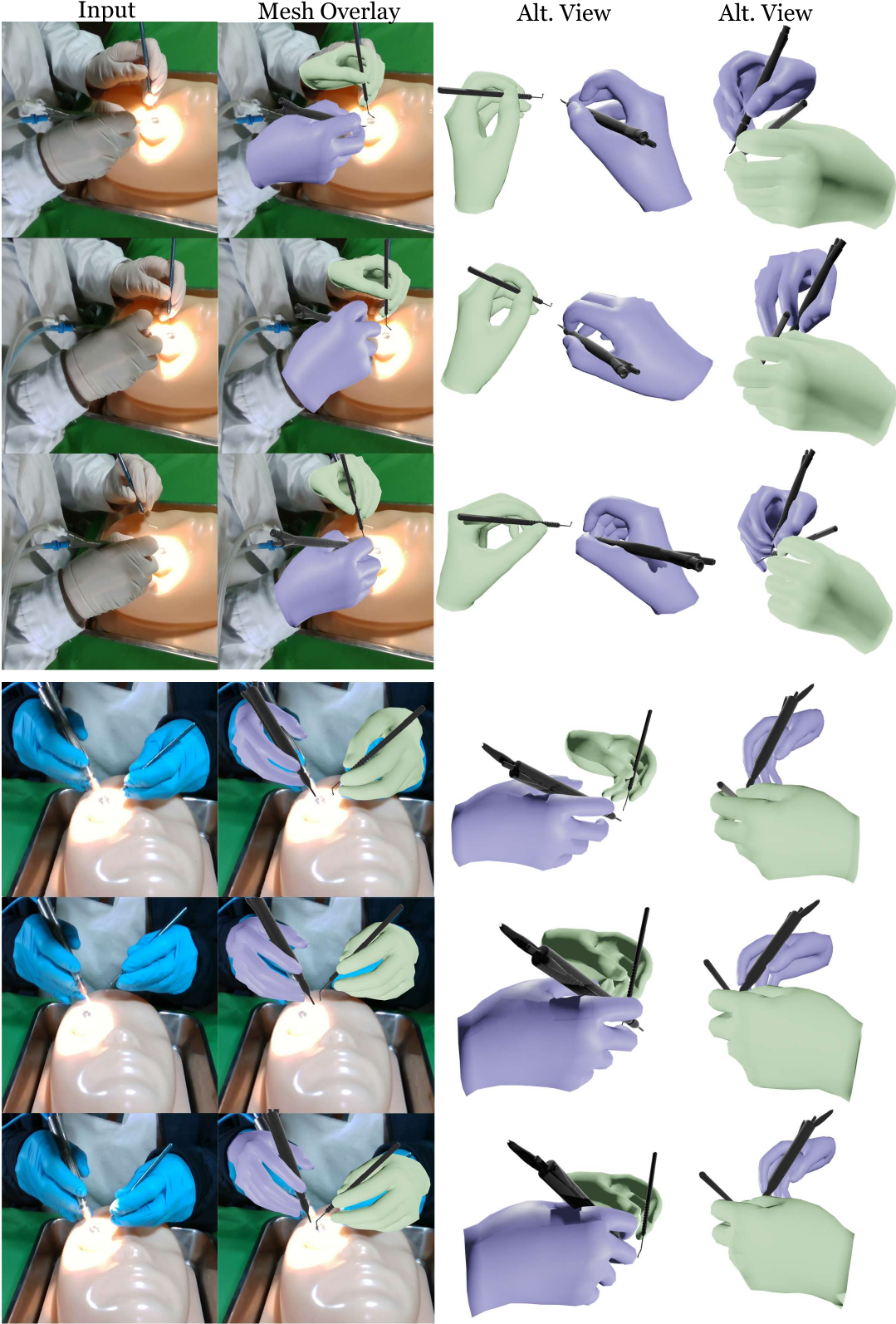}
       \caption{\textbf{Qualitative results on the hand-instrument interaction benchmark.} Each row shows a sample from the test set, with columns displaying: (1) input RGB image, (2) mesh overlay prediction, and (3)(4) for alternative view.\vspace{-4mm}}
       \label{fig:qualitative:appendix} 
\end{figure}

In this section, we provide qualitative visualizations of our model predictions on the Hand-Instrument Interaction benchmark for different phases. Each row in \cref{fig:qualitative:appendix} illustrates three temporally adjacent frames from representative video clips, capturing various surgical manipulation phases and interaction types. For each frame, we show: (1) the input RGB image, (2) the mesh overlay with predicted hand and instrument meshes, and two alternative views to highlight the spatial relationship between hands and tools. These results demonstrate that our method generates consistent, physically plausible reconstructions across frames despite visual challenges such as occlusion, rapid tool motion, and complex hand articulation. The visual continuity across time confirms that our model not only produces accurate per-frame predictions but also maintains coherent temporal behavior, which is essential for understanding fine-grained surgical actions.

\begin{figure}[th]
       \centering
       \includegraphics[width=\linewidth]{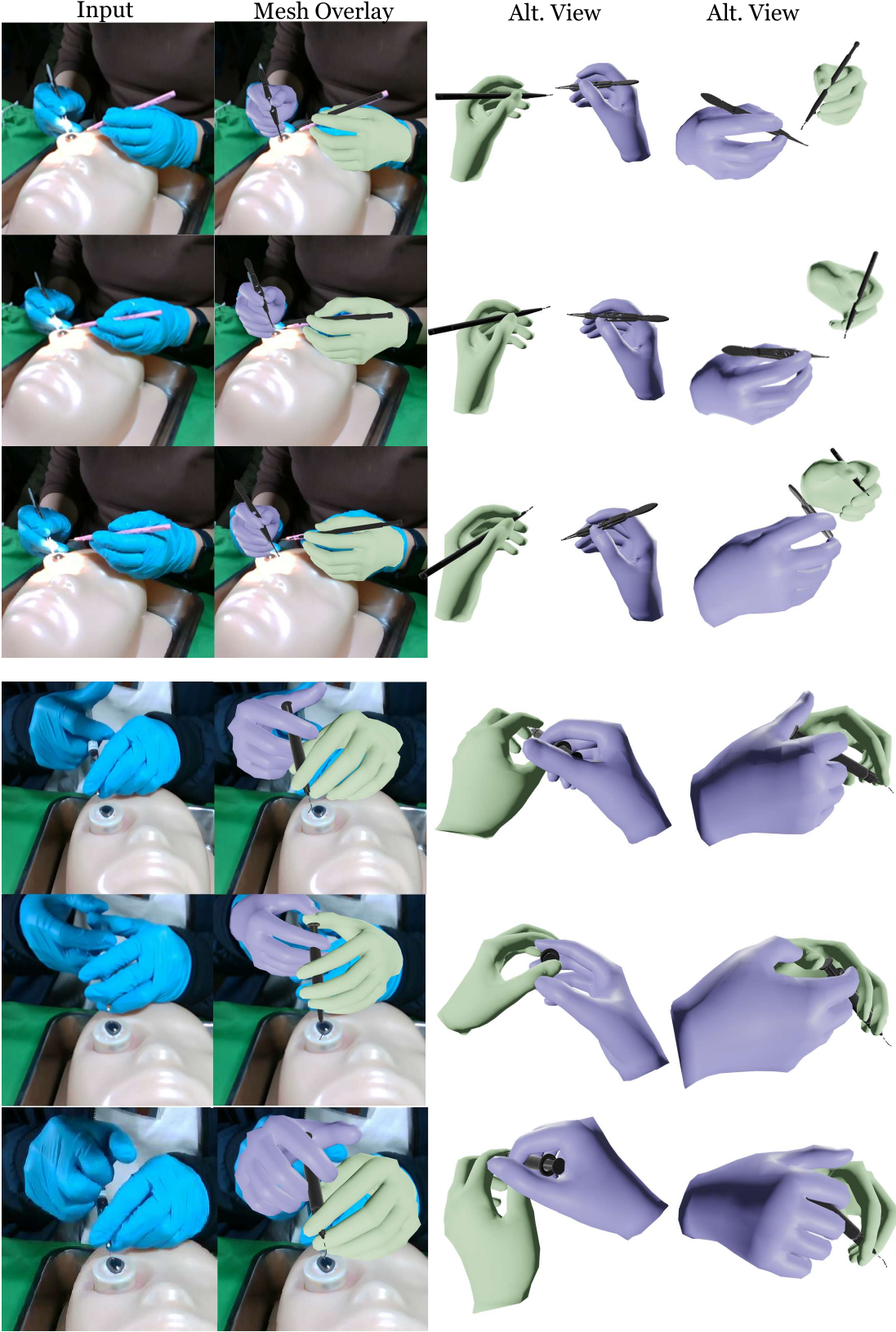}
       \caption{\textbf{Qualitative results on the hand-instrument interaction benchmark.} Each row shows a sample from the test set, with columns displaying: (1) input RGB image, (2) mesh overlay prediction, and (3)(4) for alternative view.\vspace{-4mm}}
       \label{fig:qualitative2:appendix} 
\end{figure}

\section{Discussion}
\label{dicussio:appendix}
\noindent \textbf{Limitation.}
While our dataset and method offer new insights into dynamic 3D reconstruction in ophthalmic surgery, several limitations remain. First, the data collection was conducted based on a single surgical procedure and within a single-center setting, potentially limiting the generalizability to datasets with varying surgical workflows, surgeon-specific operational habits, and illumination conditions; future work will expand to multi-center studies. Second, due to strong illumination from the surgical microscope, some instrument tips are overexposed in RGB views, affecting visibility and downstream pose estimation. Incorporating mocap-synchronized RGB capture or infrared cameras may help mitigate this issue. Third, we have not yet explored integrating the microscope view for joint reconstruction of the ocular surface, hands, and instruments, which may enable more clinically meaningful applications.

\clearpage
\begin{figure*}[t!]
\includegraphics[width=1\textwidth]{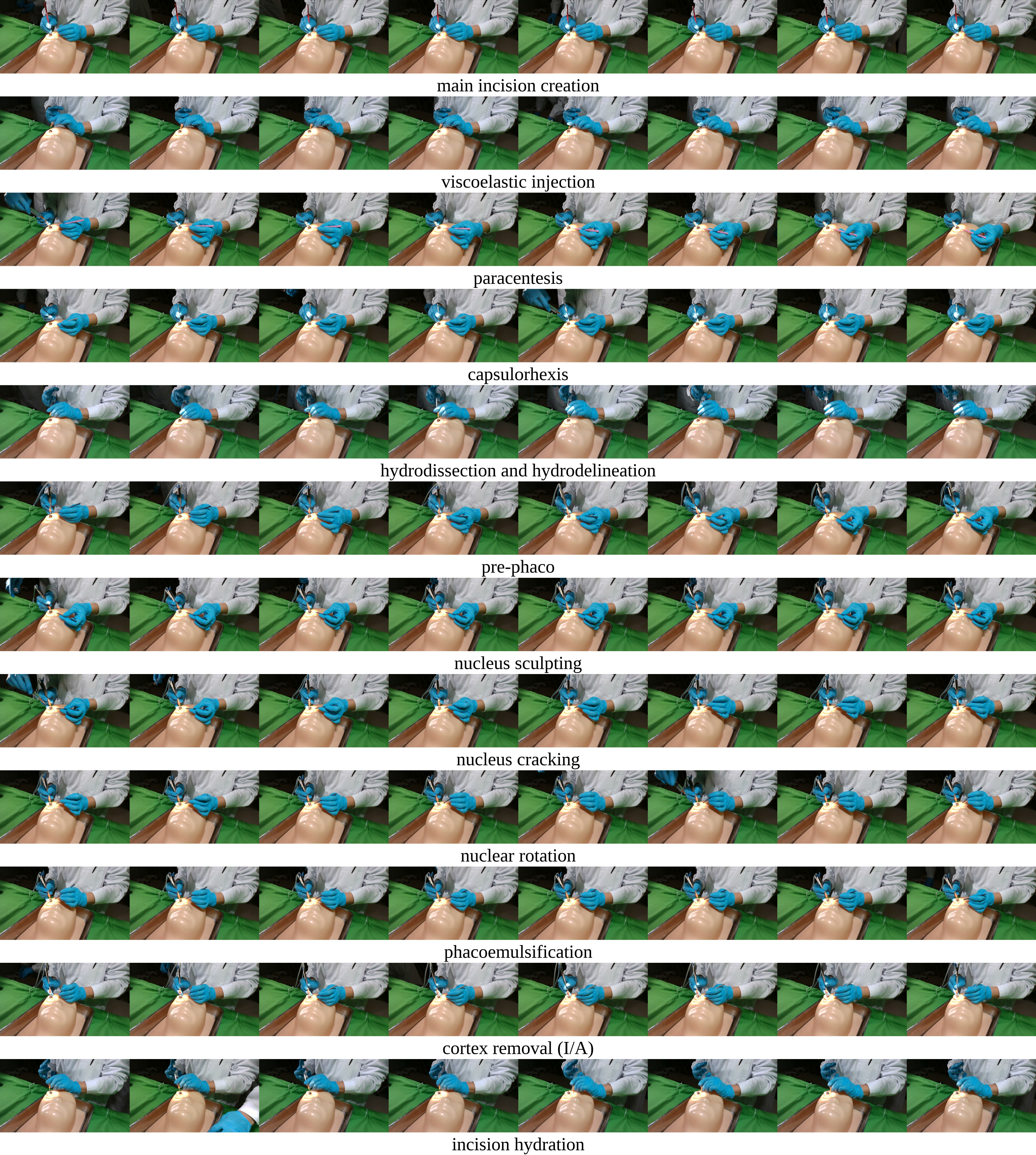}
    \centering
    \label{phase_pre1}
\end{figure*}

\begin{figure*}[t!]
\includegraphics[width=1\textwidth]{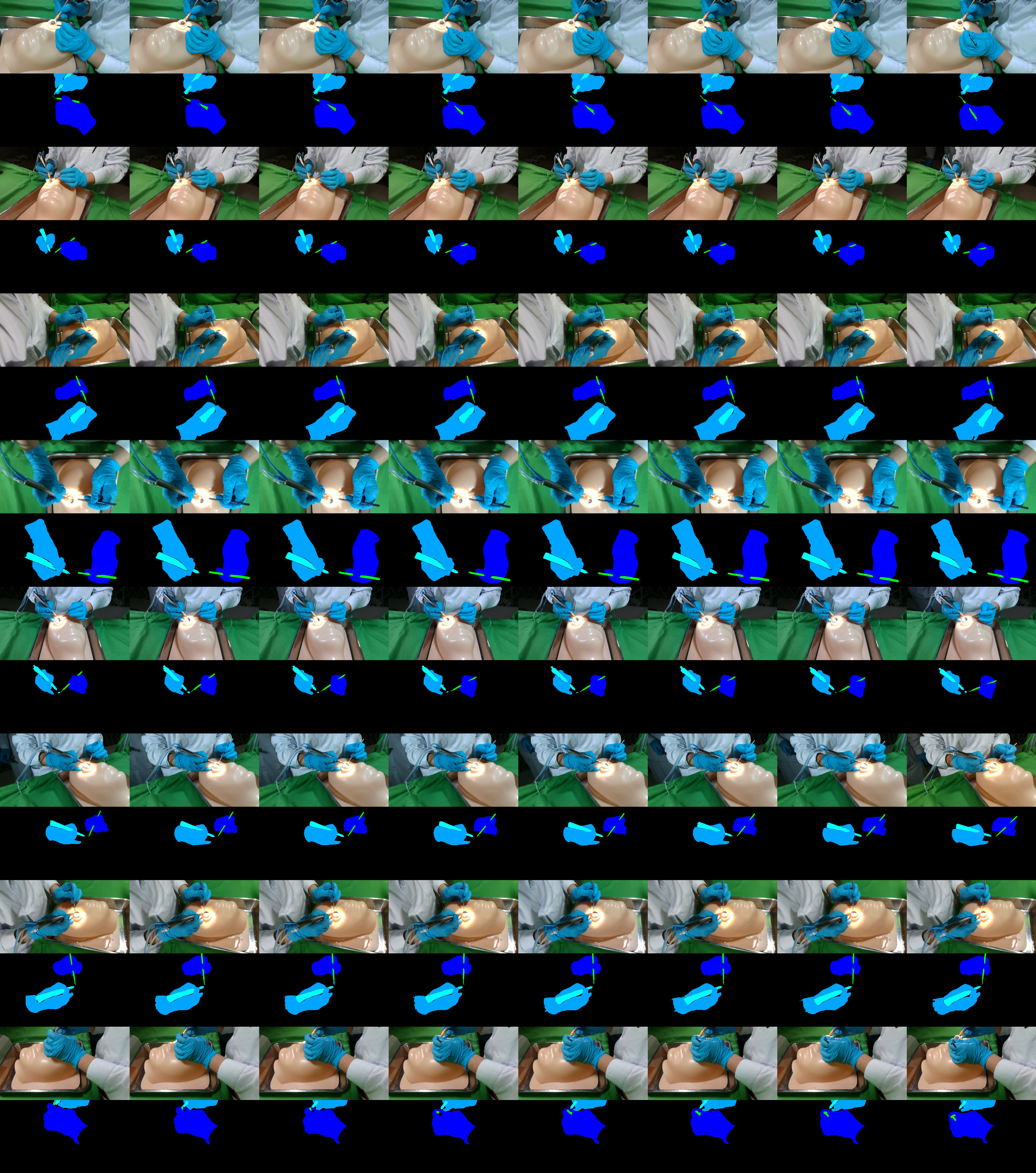}
    \centering
    \label{mask1}
\end{figure*}

\begin{figure*}[t!]
\includegraphics[width=1\textwidth]{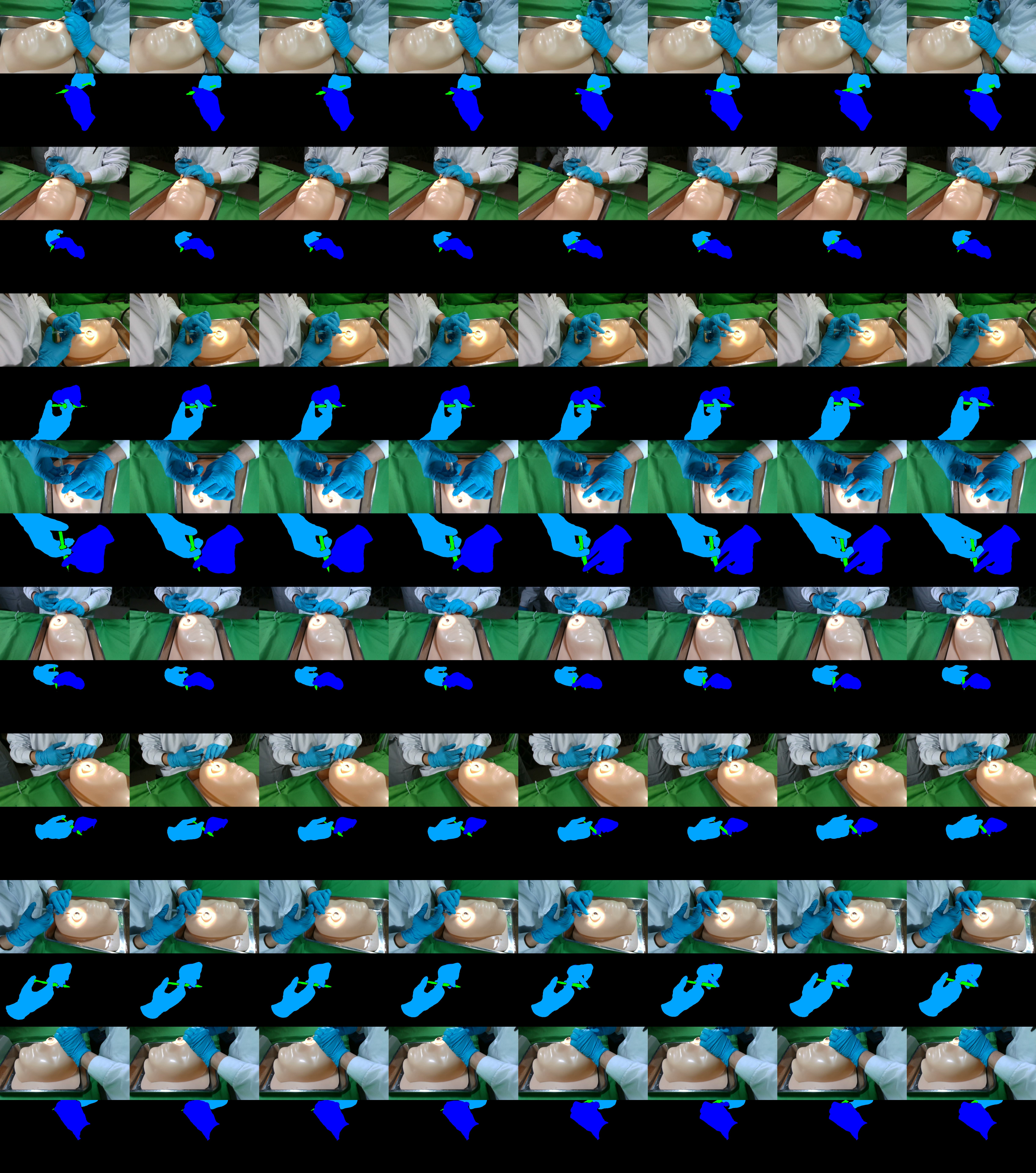}
    \centering
    \caption{Instance mask examples for phacoemulsification and viscoelastic injection.}
    \label{mask2}
\end{figure*}

\clearpage

\captionsetup[subfigure]{labelformat=empty}
\begin{figure*}[t!]
    \begin{subfigure}[t]{0.49\textwidth}
        \includegraphics[width=\textwidth]{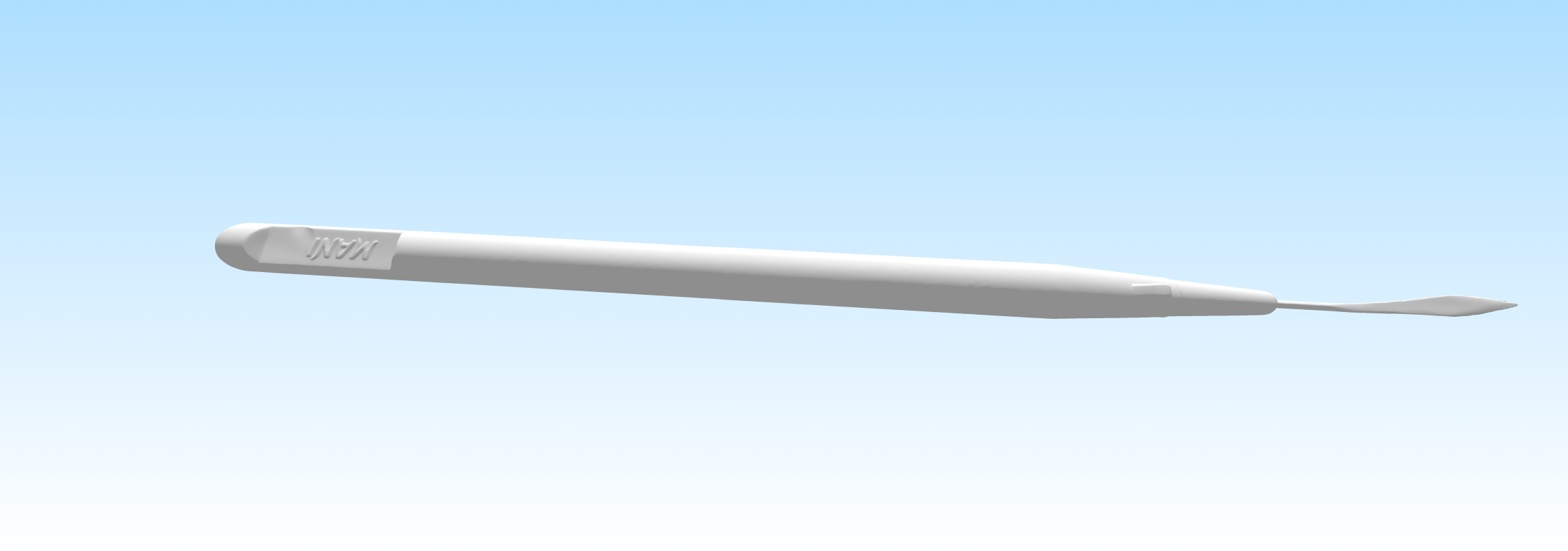}
        \caption{keratome blade}
        \label{paltform_A}
    \end{subfigure}
    \hfill
    \begin{subfigure}[t]{0.49\textwidth}
        \includegraphics[width=\textwidth]
        {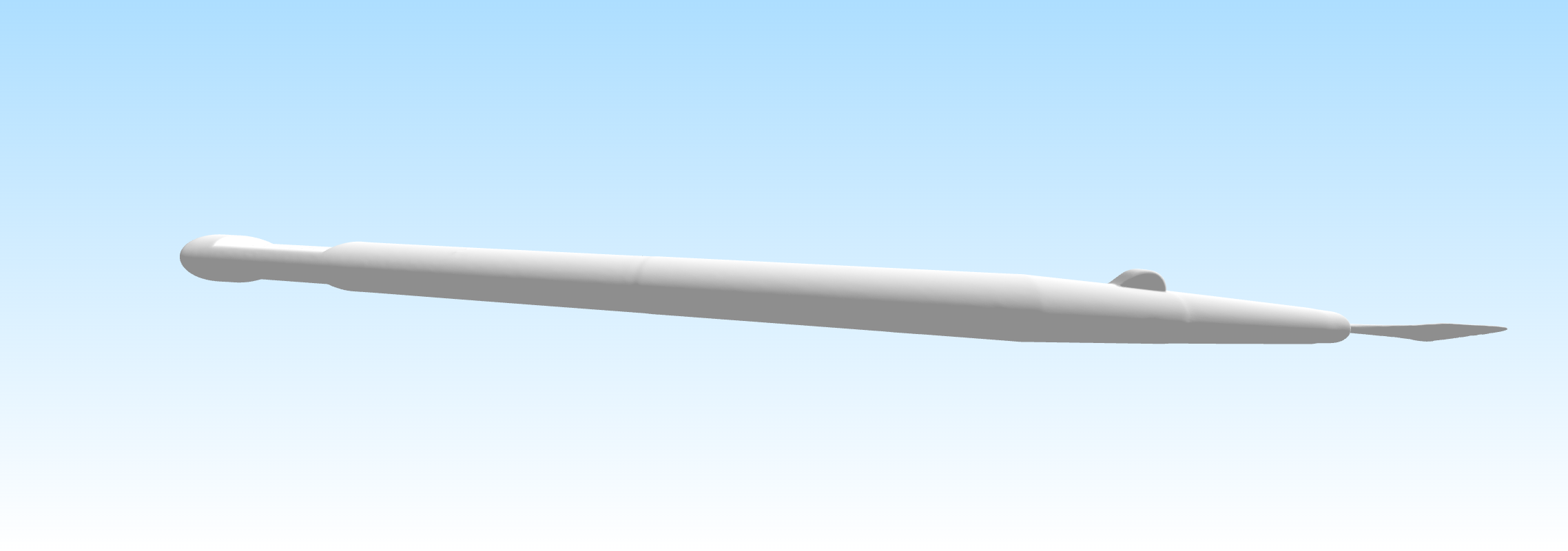}
        \caption{15° stab blade}
        \label{paltform_B}
    \end{subfigure}
    \vspace{-0.3cm}
\end{figure*}

\begin{figure*}[t!]
    \begin{subfigure}[t]{0.49\textwidth}
        \includegraphics[width=\textwidth]{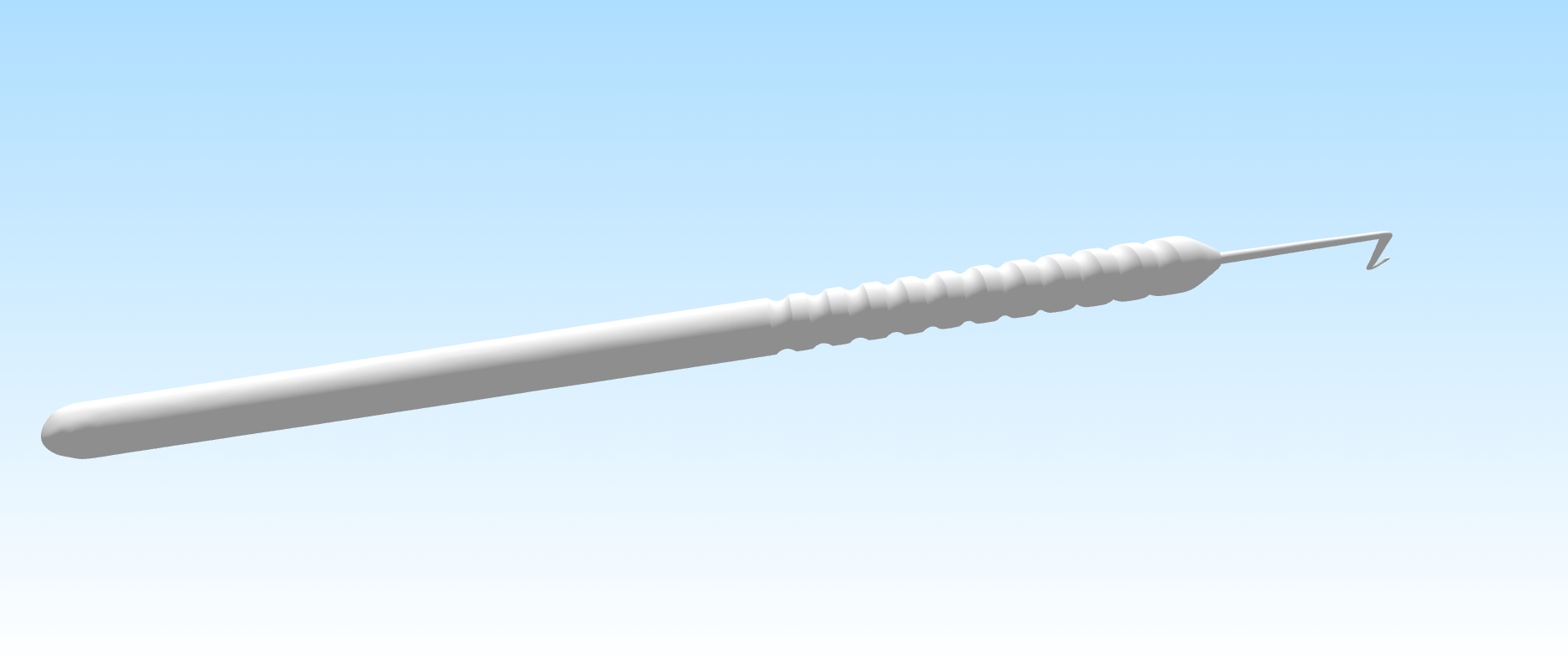}
        \vspace{-3pt}
        \includegraphics[width=\textwidth]{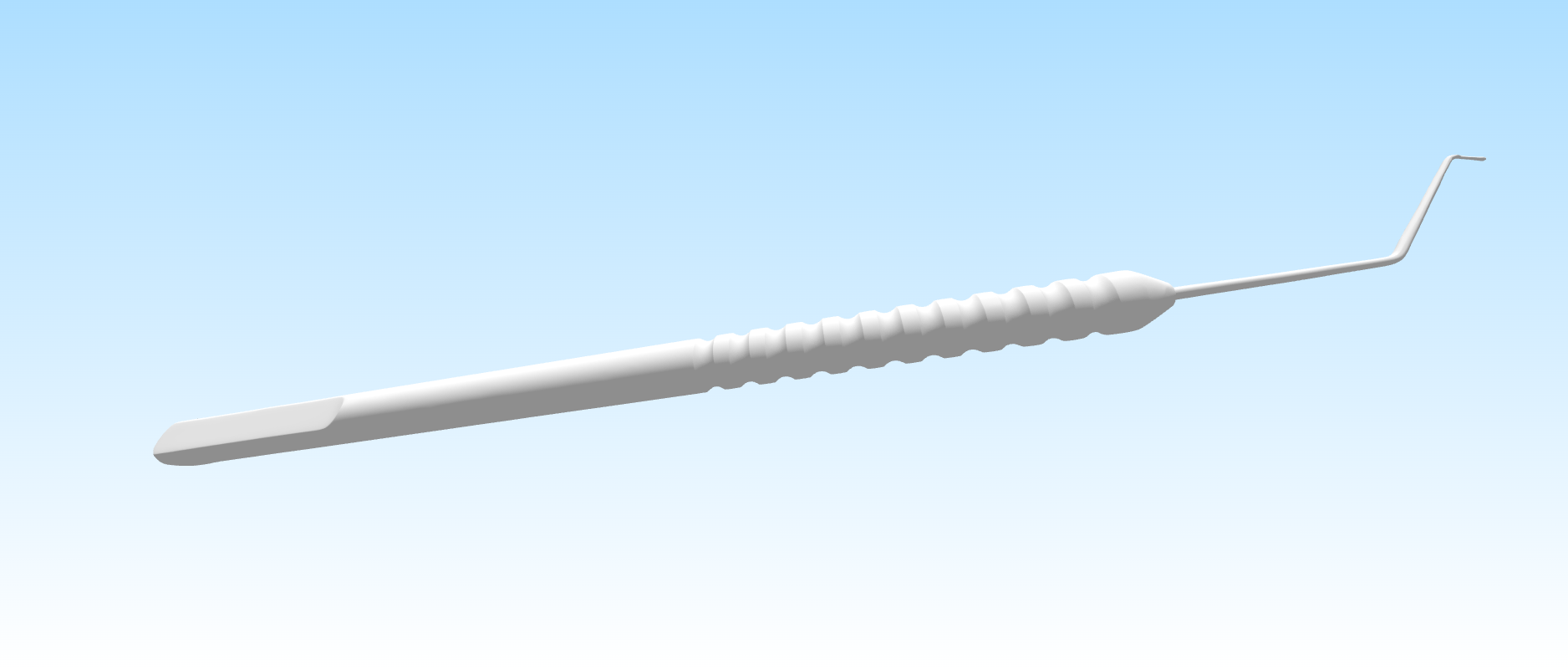}
        \caption{nucleus chopper}
        \label{paltform_C}
    \end{subfigure}
    \hfill
    \begin{subfigure}[t]{0.49\textwidth}
        \includegraphics[width=\textwidth]{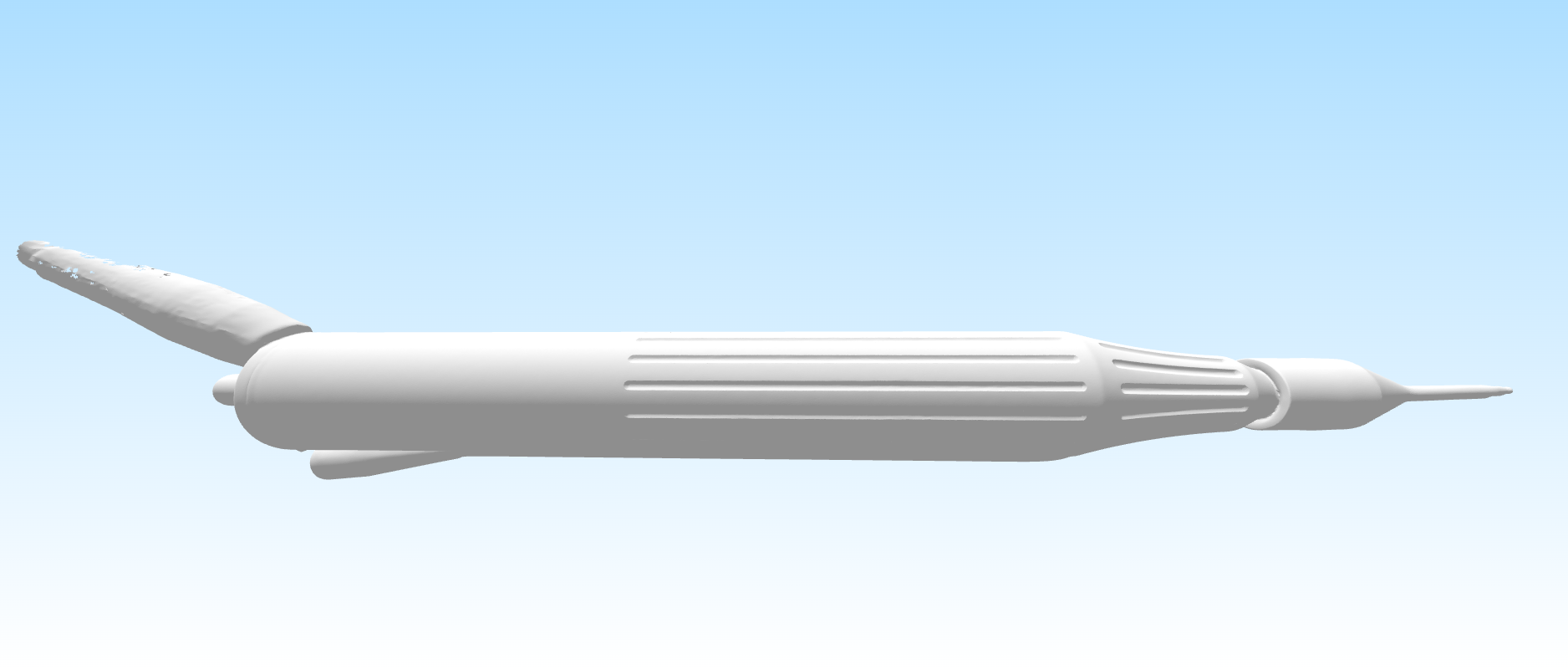}
        \vspace{10pt}
        \includegraphics[width=\textwidth]{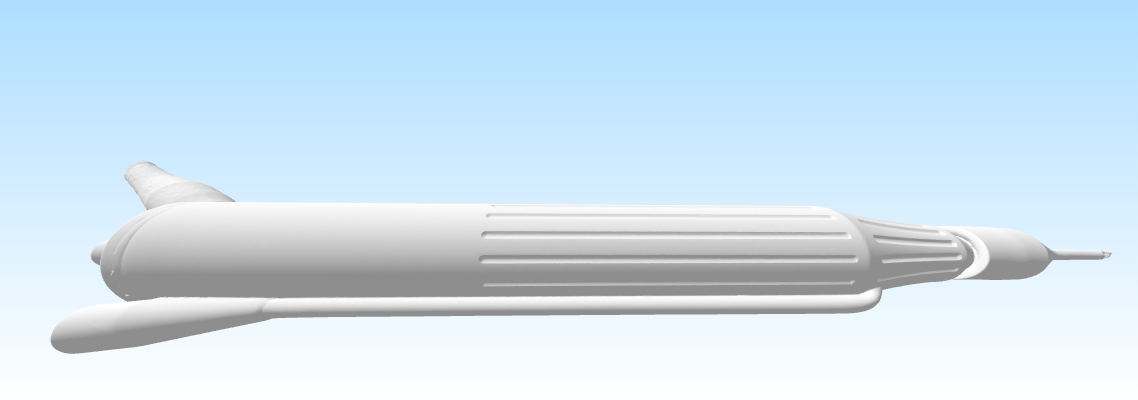}
        \caption{phacoemulsification handpiece}
        \label{paltform_D}
    \end{subfigure}

\end{figure*}

\begin{figure*}[t!]
    \begin{subfigure}[t]{0.49\textwidth}
        \includegraphics[width=\textwidth]{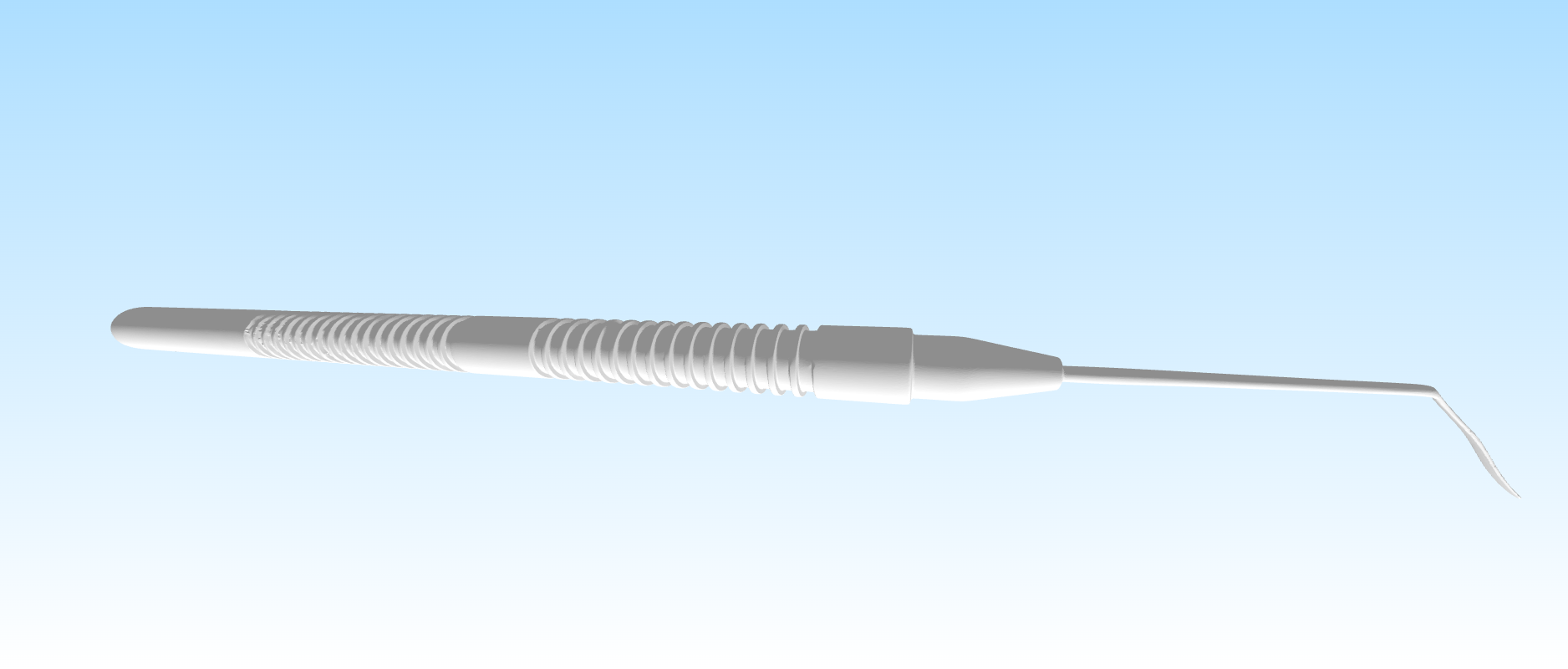}
        \caption{iris repositor}
        \label{paltform_E}
    \end{subfigure}
    \hfill
    \begin{subfigure}[t]{0.49\textwidth}
        \includegraphics[width=\textwidth]{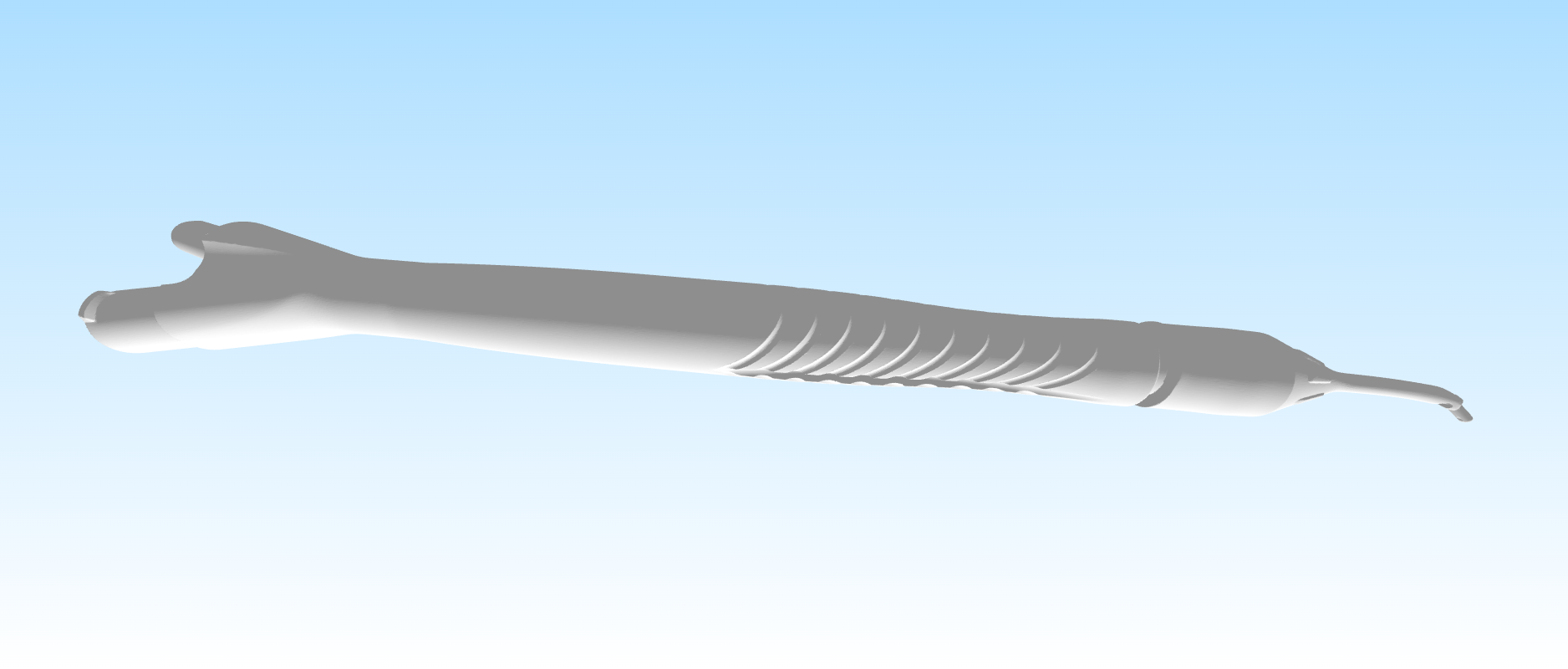}
        \caption{irrigation-aspiration handpiece.}
        \label{paltform_F}
    \end{subfigure}
    \vspace{-0.3cm}
\end{figure*}

\begin{figure*}[t!]
    \begin{subfigure}[t]{0.49\textwidth}
        \includegraphics[width=\textwidth]{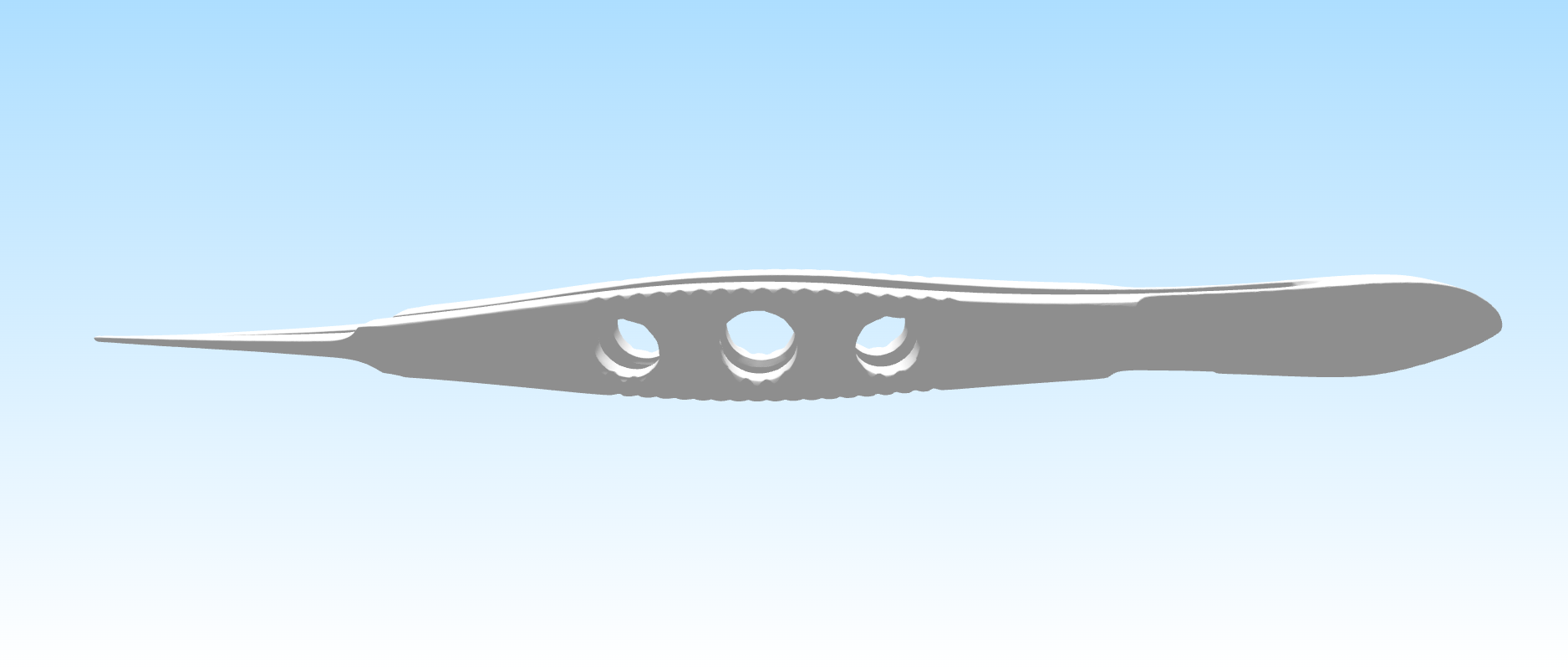}
        \vspace{-3pt}
        \includegraphics[width=\textwidth]{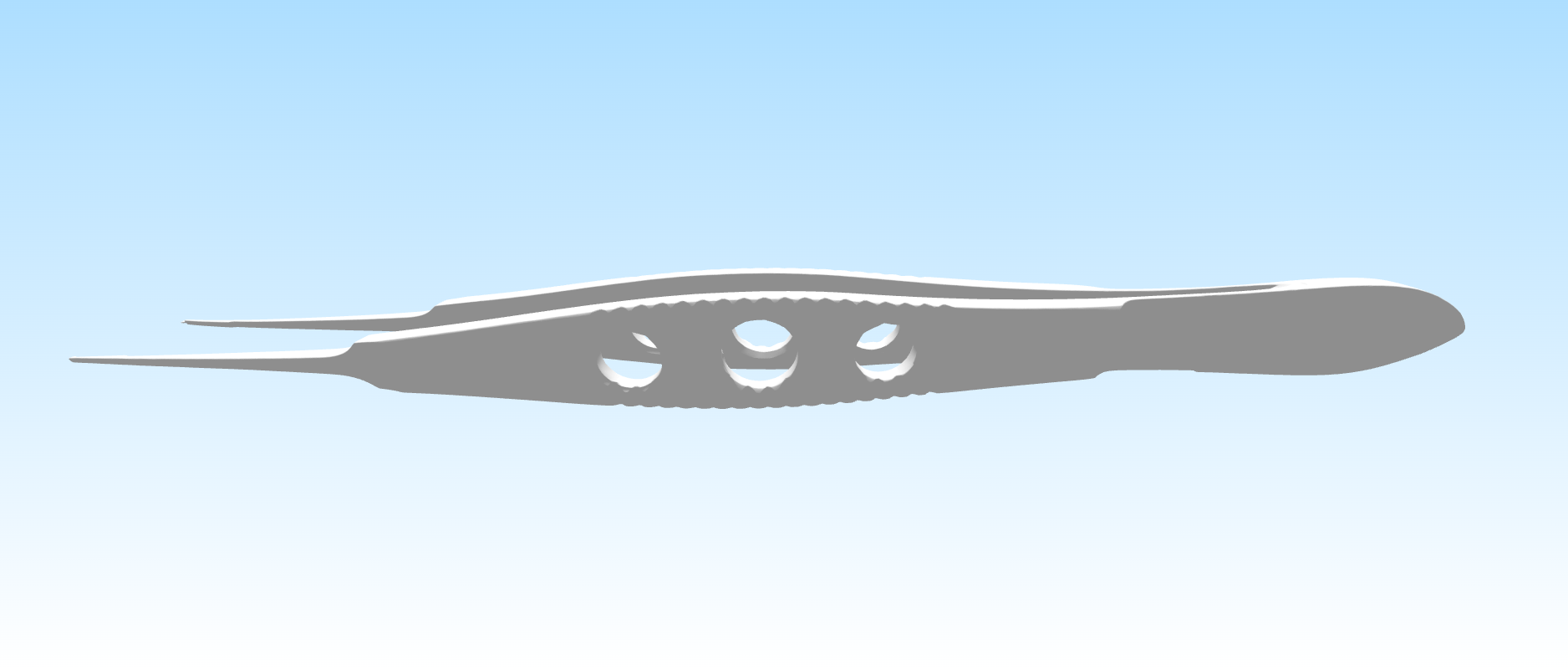}
        \caption{toothed forceps}
        \label{paltform_G}
    \end{subfigure}
    \hfill
    \begin{subfigure}[t]{0.49\textwidth}
        \includegraphics[width=\textwidth]{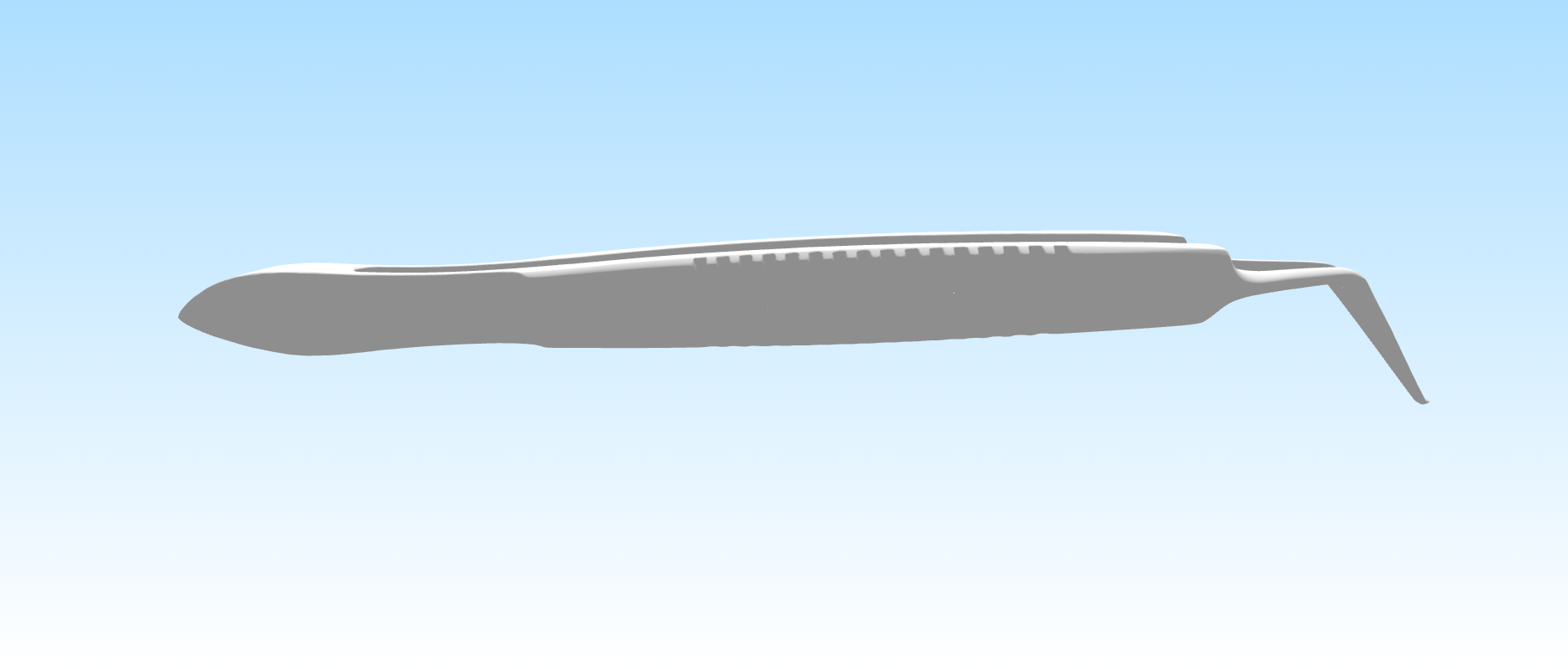}
        \vspace{-3pt}
        \includegraphics[width=\textwidth]{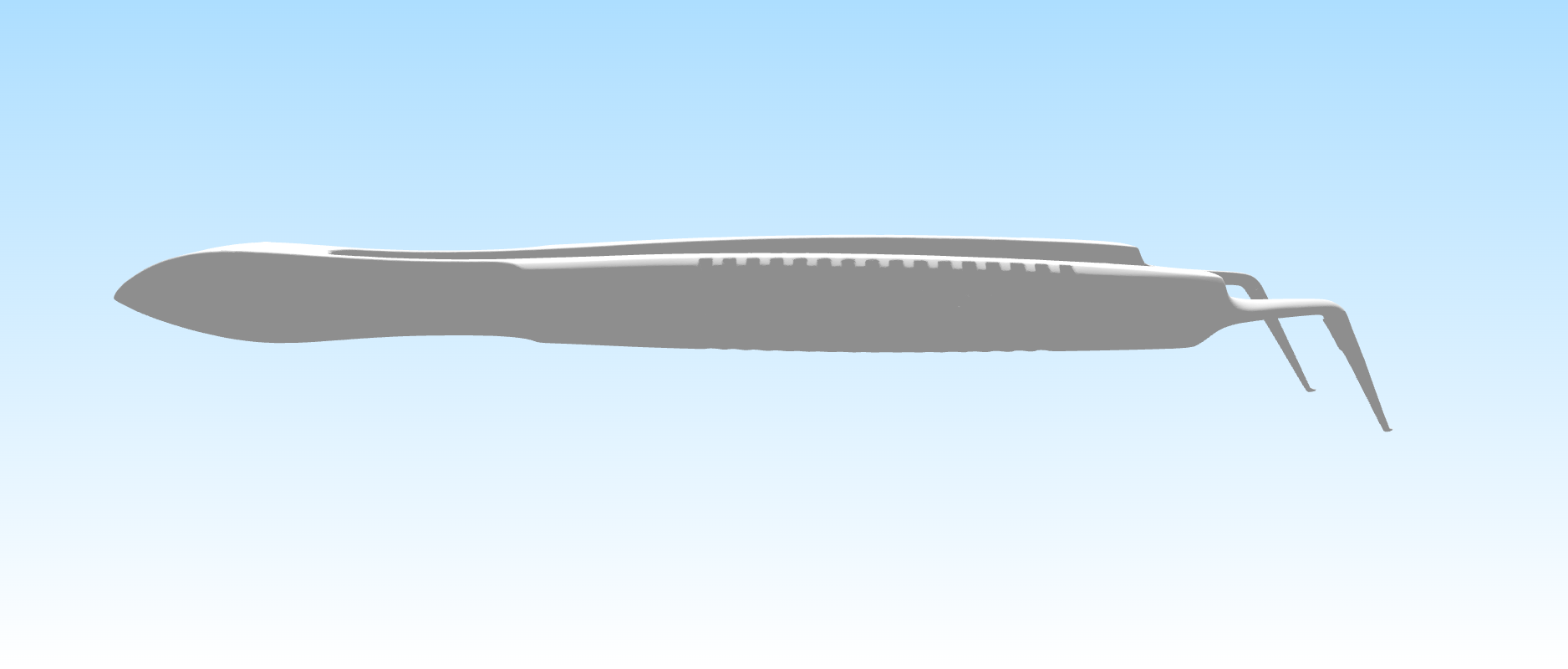}
        \caption{capsulorhexis forceps}
        \label{paltform_H}
    \end{subfigure}
    \vspace{-0.3cm}
\end{figure*}

\captionsetup[subfigure]{labelformat=parens}

\end{document}